\documentclass[sigconf]{acmart}

\AtBeginDocument{%
  \providecommand\BibTeX{{%
    \normalfont B\kern-0.5em{\scshape i\kern-0.25em b}\kern-0.8em\TeX}}}

\copyrightyear{2020}
\acmYear{2020}
\setcopyright{acmcopyright}
\acmConference[KDD '20] {26th ACM SIGKDD Conference on Knowledge Discovery and Data Mining}{August 23--27, 2020}{Virtual Event, USA}
\acmBooktitle{26th ACM SIGKDD Conference on Knowledge Discovery and Data Mining (KDD '20), August 23--27, 2020, Virtual Event, USA}
\acmPrice{15.00}
\acmDOI{10.1145/3394486.3403165}  
\acmISBN{978-1-4503-7998-4/20/08} 

\usepackage{amsmath,amsfonts,bm}

\def\eqref#1{equation~\ref{#1}}

\def\floor#1{\lfloor #1 \rfloor}
\def\1{\bm{1}}

\DeclareMathAlphabet{\mathsfit}{\encodingdefault}{\sfdefault}{m}{sl}
\SetMathAlphabet{\mathsfit}{bold}{\encodingdefault}{\sfdefault}{bx}{n}

\usepackage{graphicx}
\usepackage{multirow}
\usepackage{mathtools}
\usepackage{wrapfig}

\author{Surgan Jandial}
\authornote{Both authors contributed equally to this research.}
\authornote{contributed to project while intern at Adobe}
\affiliation{
   \institution{IIT Hyderabad} 
}
\author{Ayush Chopra}
\authornotemark[1]
\authornote{corresponding author}
\affiliation{\institution{Media and Data Science Research Lab, Adobe}}
\author{Mausoom Sarkar}
\affiliation{\institution{Media and Data Science Research Lab, Adobe}}
\author{Piyush Gupta}
\affiliation{\institution{Media and Data Science Research Lab, Adobe}}
\author{Balaji Krishnamurthy}
\affiliation{\institution{Media and Data Science Research Lab, Adobe}}
\author{Vineeth Balasubramanian}
\affiliation{\institution{IIT Hyderabad}}
\begin{abstract}
Deep neural networks (DNNs) are powerful learning machines that have enabled breakthroughs in several domains. In this work, we introduce a new retrospective loss to improve the training of deep neural network models by utilizing the prior experience available in past model states during training. Minimizing the retrospective loss, along with the task-specific loss, pushes the parameter state at the current training step towards the optimal parameter state while pulling it away from the parameter state at a previous training step. Although a simple idea, we analyze the method as well as conduct comprehensive sets of experiments across domains - images, speech, text and graphs - to show that the proposed loss results in improved performance across input domains, tasks, and architectures.
\end{abstract}


\begin{document}
\fancyhead{}
\title{Retrospective Loss: Looking Back to Improve Training of Deep Neural Networks}

\maketitle

\section{Introduction}

Deep neural network (DNN) models have enabled breakthroughs in varied fields such as computer vision, speech recognition, natural language understanding and reinforcement learning  in recent years. In addition to extending their success to newer application domains, the last few years have also seen significant efforts in improving the training of DNN models and improve generalization performance through data augmentation, regularization methods and various new training strategies \citep{random-data-aug} , \citep{noise-regularization},  \citep{dsd-training}. In this work, we introduce a new perspective to training DNN models - retrospection - which seeks to improve DNN training by utilizing prior experiences (past model states) of the DNN during training itself.

Humans are efficient learners with the ability to quickly understand and process diverse ideas. A key aspect of human intelligence that enables efficient learning is the capability to actively reference past experiences, including past versions of one's own personality, to continually improve and adapt oneself to achieve better performance at tasks in the future. One would ideally like artificial learning agents that we create to also learn and be inspired by facets of human learning, including the ability to learn from the past and adapt quickly. While DNN models do learn from past data during training, we focus on a different aspect - learning from their own past model states during training (equivalent to humans learning from behavior of previous versions of themselves) - in this work. 

\begin{figure}[t]
    \centering
    \includegraphics[scale=0.25]{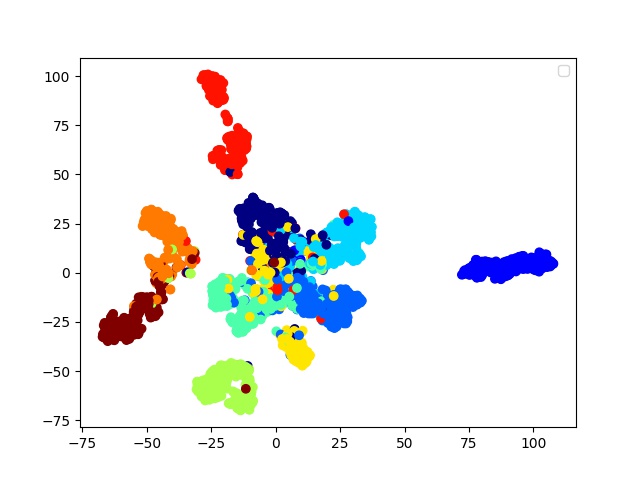}
    \includegraphics[scale=0.25]{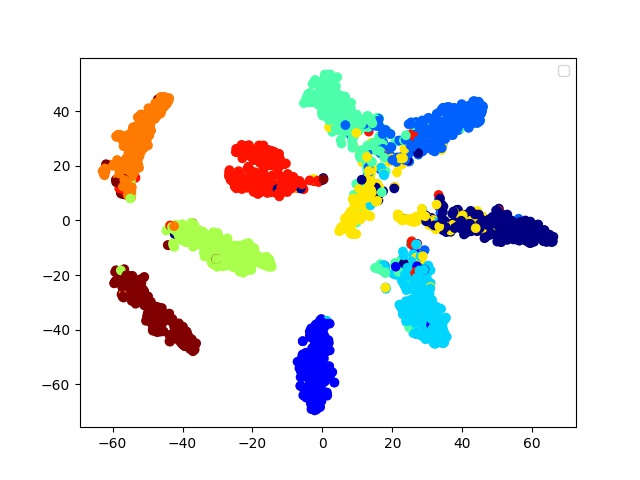}
    \vspace{-2pt}
    \caption{Illustration of an outcome of the proposed retrospective loss. Figure shows t-SNE plots of representations of different classes learned by training the LeNet architecture on FMNIST dataset \textit{(left)} without and \textit{(right)} with retrospective loss. The inclusion of the proposed loss term along with standard cross-entropy loss in this simple case significantly improves discriminability of representations.}
    \vspace{-8pt}
    \label{fig_tsne_plots_lenet}
\end{figure}
We introduce a simple new idea, \textit{retrospective loss}, that utilizes prior training experiences in the form of DNN model states during training to guide weight updates and improve DNN training performance. The proposed loss seeks to ensure that the predictions at a particular training step are more similar to the ground truth than to the predictions from a previous training step (which has relatively poorer performance). As training proceeds, minimizing this loss along with the task-specific loss, encourages the network parameters to move towards the optimal parameter state by pushing the training into tighter spaces around the optimum. The proposed retrospective loss is easy to implement, and we empirically show through a comprehensive set of experiments across domains and corresponding ablation studies that it works well across input domains (images, speech, text and graphs), multiple tasks, as well as network architectures. Figure \ref{fig_tsne_plots_lenet} shows an illustrative example of the use of retrospective loss in the simple case of LeNet training on the FMNIST dataset, which results in significantly improved representations (described further in Sec \ref{sec_expts}).

\begin{figure*}[t]
\centering
\includegraphics[width=12cm,keepaspectratio]{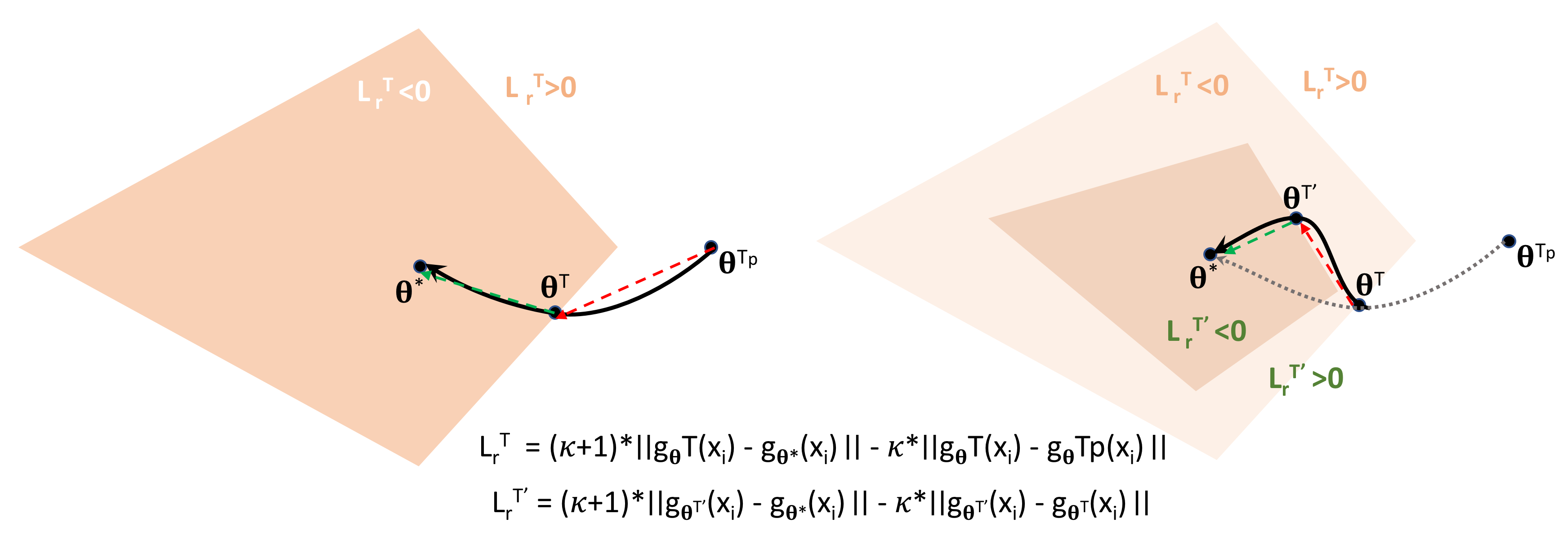}
\vspace{-9pt}
\caption{Geometric intuition of the working of the proposed retrospective loss. The figures show polytopes in the weight parameter space. \textit{(Left)} For all $\theta^i$ inside the shown colored polytope, the retrospective loss is negative and is positive outside. Our objective is to push parameters of the current $\theta^T$ further inside this polygon close to $\theta^*$; \textit{(Right)} In a future time step $T' > T$, by design of the retrospective loss, the polytope region shrinks and our objective at this time step is to push parameters to a near-optimal region around $\theta^*$.}
\vspace{-7pt}
\label{fig_geom_intuition}
\end{figure*}
The key contributions of our work can be summarized as follows: (i) We propose a new simple, easy to implement \textit{retrospective loss} that is based on looking back at the trajectory of gradient descent and providing an earlier parameter state as guidance for further learning (Sec \ref{sec_methodology}); (ii) We exhaustively experiment across domains - images, speech, text and graphs, as well as on a range of tasks including classification,few-shot learning and GANs, and consistently beat state-of-the-art methods on benchmark datasets with the addition of this loss term during training (Sec \ref{sec_expts}); (iii) We analyze and explain the intuition and the reasoning behind why it works (Sec \ref{sec_methodology}), as well as conduct ablation studies to study the impact of various choices in Sec \ref{sec_analysis}. To the best of our knowledge, this is a first such effort in its perspective, and our empirical studies show consistent improvement in performance across tasks in our multiple trials, demonstrating potential for practical use in real-world applications across domains.

\section{Related Work}
\label{sec_related_work}

The proposed loss leverages parameter states from previous training steps as guidance to compute the current weight update. While there is no explicit effort that implements this idea when training DNN models, one could find similarities with efforts in optimization that utilize information from past training steps for current weight updates. Techniques such as SVRG \citep{svrg}, SARAH\citep{sarah}, and ProxSARAH \citep{ProxSARAH} use gradients from earlier training steps to predict better weight updates. These methods are very different in their objectives, and most often seek to reduce variance in the stochastic gradient descent update. Other optimization methods like Momentum, Adam,  Nesterov Momentum accumulate past gradients to accelerate weight updates in the right direction in order to achieve faster convergence. While these methods leverage the most recent updates to provider a stronger gradient direction, in contrast, our work seeks to leverage the relatively poorer performance of past models states during training. The proposed method can be used to improve performance when used with different optimizer configurations (including Adam or momentum), as shown in our results.

Another genre of efforts could be traced to reinforcement learning (RL), where techniques involve optimizing using moving targets. In such settings, methods for Q-learning and policy gradients benefit from using a guidance network during training. The DQN algorithm proposed by \citep{mnih-nature} uses an additional target network for Q-value updates, where parameters are updated by copying from the additional network at discrete steps. Double Q-learning \citep{double-qn} learns two Q functions, where each Q-function is updated with a value for the next state from the other Q-function. Policy gradient methods such as TRPO \citep{TRPO}, PPO \citep{PPO} use a KL-divergence objective during training that constrains the loss to ensure deviation from a previously learned policy is small. In these techniques, leveraging a guidance during training results in improved convergence and sample efficiency. We note that these efforts, while similar in a sense, are constrained to the RL setting, and the proposed work is the first such effort in supervised learning to the best of our knowledge. Further, the objective in the RL setting is to control divergence from a guidance step to better handle moving targets, which is fundamentally different from the objectives in supervised learning. We note additionally that while the proposed retrospective loss is intended to be added to any task-specific loss, it is not a regularizer (not intended to overcome overfitting), but to improve efficiency and efficacy of DNN model training.

\section{Methodology}
\label{sec_methodology}

We now present the formulation and analysis of the proposed retrospective loss. For ease of understanding, we begin by introducing the notations and the loss itself. We subsequently present the conceptual formulation of the loss and analyze how it works later in this section.\\
\vspace{-6pt}

\noindent \textbf{Notations.} Given a dataset $\{(\textbf{x}_i,y_i):i=1,\cdots,m\}$ of $m$ labeled training samples, we consider a neural network, $g: \mathbb{R}^n \rightarrow \mathbb{R}^d$, parametrized by its weights $\theta$ where each $\textbf{x}_i \in \mathbb{R}^n$ and $d$ represents the number of classes in classification problems and the dimensionality of output in regression problems. Let the optimal parameters of the neural network be given by $\theta^*$, i.e. $g_{\theta^{*}}(\textbf{x}_i) = y_i$ for all $i=1,\cdots,m$. At a particular time step $T$ during training, the neural network parameters are given by $\theta^T$. For convenience and brevity, we ignore the input and write $g_{\theta}(\textbf{x}_i)$ as simply $g_{\theta}$ at certain parts of the paper.\\
\vspace{-6pt}

\noindent \textbf{Proposed Retrospective Loss.} The proposed retrospective loss is designed  to leverage past model states during training, and cue the network to be closer to the optimal model parameters than a state in the past. In other words, minimizing it with respect to $\theta$ during training seeks to constrain the model parameter state at each time step $\theta^{T}$ to be closer to $\theta^*$ than a model parameter state from a past time step, $\theta^{T_p}$. Given an input data-label pair $(\textbf{x}_i, y_i)$, the retrospective loss at time step $T$ is given by:
\begin{equation}
\label{eqn_retrospective_loss}
    \mathcal{L}_{retrospective} = (\kappa+1) ||g_{\theta^T}(\textbf{x}_i) - y_i|| - \kappa ||g_{\theta^T}(\textbf{x}_i) - g_{\theta^{T_p}}(\textbf{x}_i)||
\end{equation}
\noindent The $\kappa$-based scaling co-efficients are included with a purpose, which is substantiated in the analysis later in this section. 

Adding this loss term to an existing supervised learning task loss provides for efficient training, which is validated in our experiments across the domains of images, speech, text and graphs in Section \ref{sec_expts}. The retrospective loss is introduced to the training objective following a warm-up period ($I_w$) wherein the neural network function can be considered stable for use of such retrospective updates. The training objective at any training step $T$ with the retrospective loss is hence $\mathcal{L} = \mathcal{L}_{task} + \mathcal{L}_{retrospective}$, where $\mathcal{L}_{task}$ is the task-specific training objective (such as cross-entropy loss for classification, mean-squared error for regression or any other such loss for that matter).\\ 
\vspace{-6pt}

\noindent \textbf{Intuition.} Figure \ref{fig_geom_intuition} illustrates the geometric intuition of the working of the retrospective loss. 
By design (Eqn \ref{eqn_retrospective_loss}), $\mathcal{L}_{retrospective}$ is negative when the current model state, $g_{\theta^T}$ is farther away from the retrospective step, $g_{\theta^{T_p}}$, than the optimal solution $g_{\theta^*}$. 
One could view the loss term as dividing the parameter space into two regions: a polytope around the optimal $\theta^*$ where $\mathcal{L}_{retrospective} < 0$, and the region outside the polytope where $\mathcal{L}_{retrospective} > 0$. 
Minimizing retrospective loss pushes the network towards parameters further inside the polytope, thus helping speed up the training process. As shown on the right subfigure in Figure \ref{fig_geom_intuition}, the polytope shrinks over time, since the retrospective support, $T_p$, is also updated to more recent parameter states. 
This helps further push the parameters into a near-optimal region around $\theta^*$. The loss term helps in improved solution in most cases, and faster training in certain cases, as shown in our extensive empirical studies in Section \ref{sec_expts}. Algorithm 1 summarizes the methodology.\\
\vspace{-6pt}

\begin{figure}
\centering
    \includegraphics[width=0.5\textwidth]{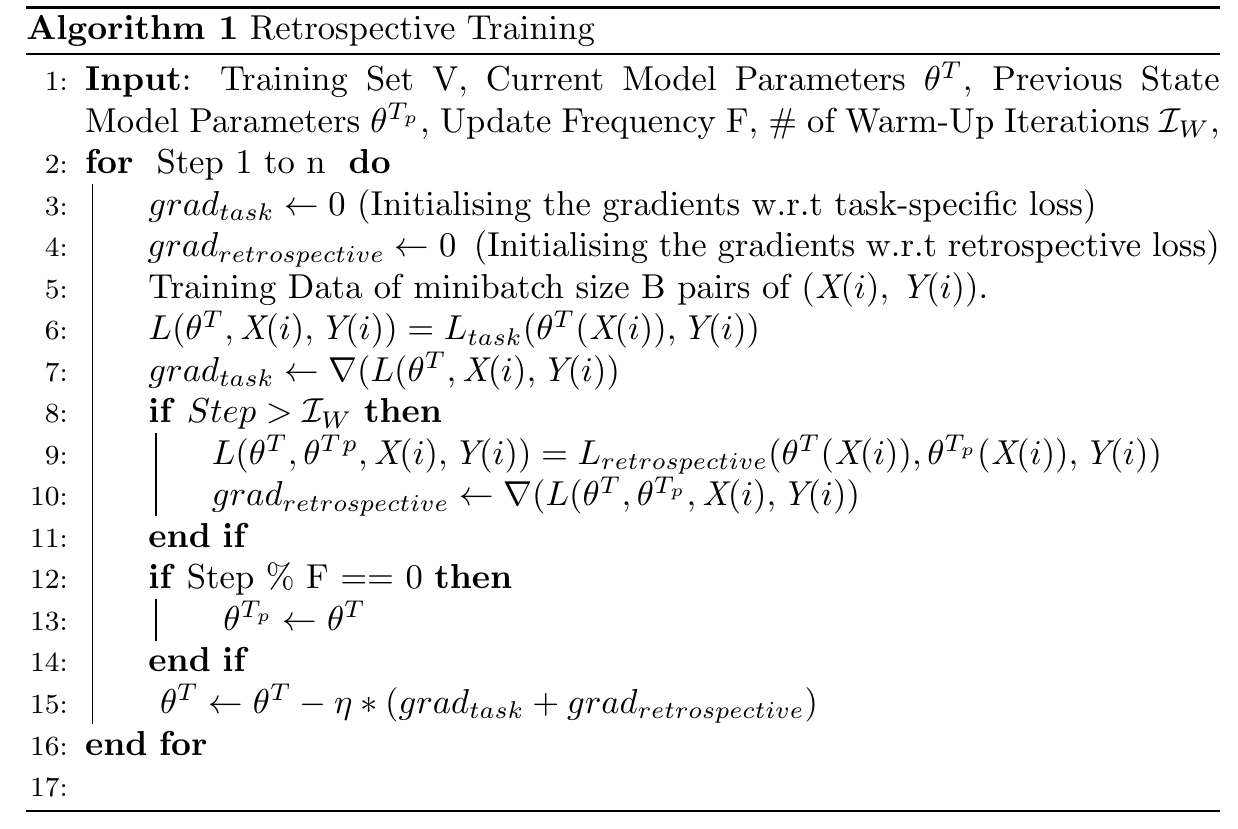}
\vspace{-25pt}
\end{figure}

\noindent \textbf{Analysis.} We begin the analysis of the proposed loss by formalizing a key property of consistency, albeit well-known, that any loss term added solely to speed up the optimization process, when minimizing an objective function, ought to satisfy. We state this in the context of neural networks here.\\
\vspace{-6pt}

\noindent \textit{\textbf{Consistency of Loss Terms for Neural Network Models.} Let $g_{\theta}$ represent the neural network with weights $\theta$, and let $\theta^*$ be the optimal weights that minimize a given loss function, $\mathcal{L}_{task}$, i.e. $\theta^* = \arg\min_{\theta}\big(\mathcal{L}_{task}(g_{\theta})\big)$. Then, the minimum for any new loss function, $\mathcal{L}_{task}$ + $\mathcal{L}_{add}$, where $\mathcal{L}_{add}$ is included solely for the purpose of speeding up optimization, ought to be maintained at $\theta^*$. i.e. $\theta^* = \arg\min_{\theta}\big(\mathcal{L}_{task}(g_{\theta}) + \mathcal{L}_{add}(g_{\theta})\big)$.}\\
\vspace{-6pt}

\begin{figure}[t]
\includegraphics[width=\linewidth]{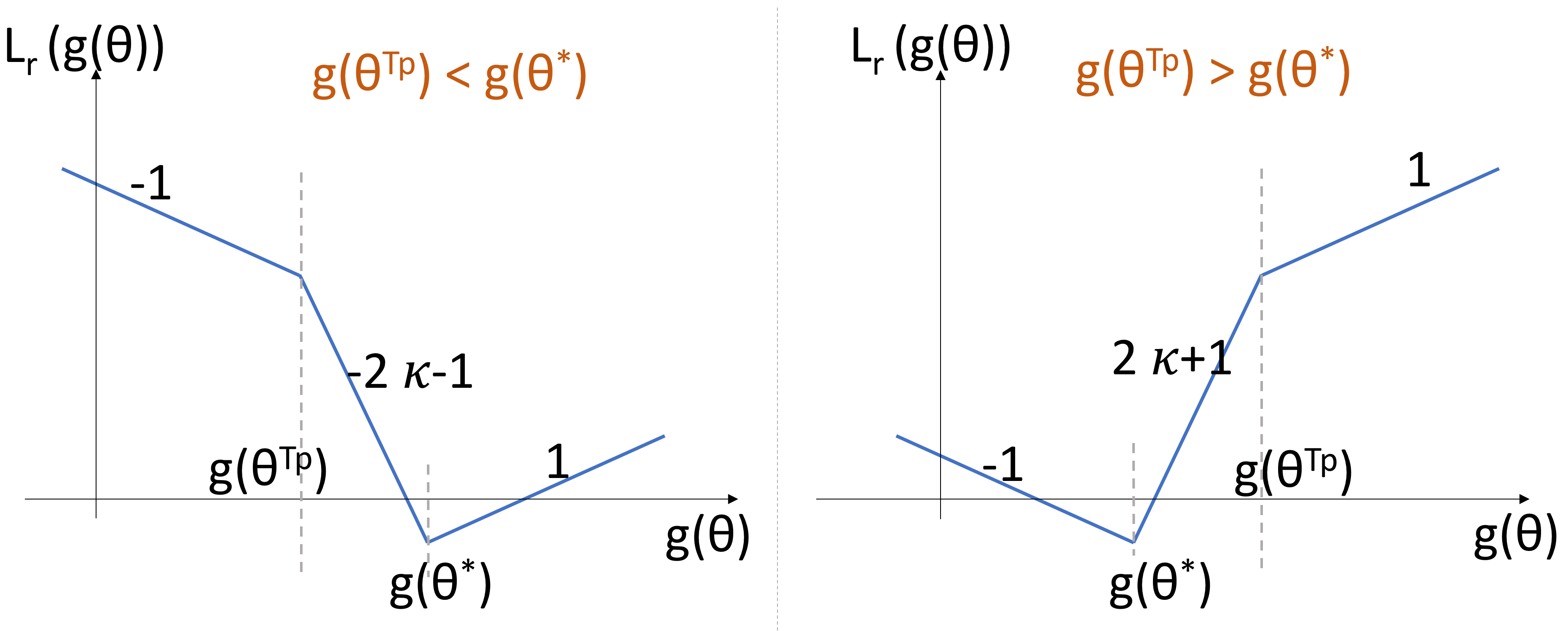}
\vspace{-9pt}
\caption{Gradient of the proposed retrospective loss. Minimum can be seen to be at $g(\theta^*)$ in the $L_1$-norm case.}
\vspace{-7pt}
\label{fig_gradient_intuition}
\end{figure}

We now study the consistency of the retrospective loss. To this end, consider the total loss, which is a sum of the task-specific loss and the retrospective loss: 
\begin{equation}
\label{total_loss}
\mathcal{L} = \mathcal{L}_{task} + (\kappa+1) ||g_{\theta^T}(\textbf{x}_i) - y_i|| - \kappa||g_{\theta ^T}(\textbf{x}_i) - g_{\theta^{T_p}}(\textbf{x}_i)||
\end{equation}
\noindent The gradient of Eqn \ref{total_loss} w.r.t. $\theta$ is given by:
\begin{equation}
    \begin{split}
    \frac {\partial}{\partial\theta} \mathcal{L} = \bigg( \frac {\partial}{\partial g(\theta)} \mathcal{L}_{task}  + 
    (\kappa+1) \frac {\partial}{\partial g(\theta)}||g_{\theta^T}(\textbf{x}_i) - y_i|| - \\ 
    \kappa \frac {\partial}{\partial g(\theta)}||g_{\theta^T}(\textbf{x}_i) -  g_{\theta^{T_p}}(\textbf{x}_i)|| \bigg)\frac {\partial g(\theta)}{\partial\theta}
    \end{split}
\end{equation}
\noindent The additional term in the gradient is hence:
\begin{equation}
\label{retro_grad}
    \bigg((\kappa+1) \frac {\partial}{\partial g(\theta)}||g_{\theta^T} - g_{\theta^*}|| - 
    \kappa \frac {\partial}{\partial g(\theta)}||g_{\theta^T} -  g_{\theta^{T_p}}|| \bigg)\frac {\partial g(\theta)}{\partial\theta}
\end{equation}
\noindent where $y_i$ is replaced with $g_{\theta^*}$ and $\textbf{x}_i$ is removed for brevity. Considering the inside term in Eqn \ref{retro_grad} and $L_1$-norm as the choice of norm, we get:
\begin{equation}
    (\kappa+1) * sgn(g_{\theta^T} - g_{\theta^*}) - \\ 
    \kappa*sgn(g_{\theta^T} -  g_{\theta^{T_p}}) 
\end{equation}
\noindent The additional contribution to the gradient due to the retrospective loss when
$g_{\theta^{T_p}} < g_{\theta^*}$ is then given by (see Fig \ref{fig_gradient_intuition}, left subfigure, note that this figure is a plot between $g_{\theta}$ on the $x$-axis and $\mathcal{L}_{g_{\theta}}$ on the $y$-axis):
\begin{equation}
\vspace{-3pt}
\label{retro_l2_grad}
    \begin{cases} 
        -1 & for\  g_{\theta^T} < g_{\theta^{T_p}} \\
        -2\kappa-1 & for\  g_{\theta^{T_p}}< g_{\theta^T} < g_{\theta^*}  \\
        1 & for\ g_{\theta^T} > g_{\theta^*} 
    \end{cases}
\end{equation}
\noindent and in case of $g_{\theta^{T_p}} > g_{\theta^*}$ is (see Fig \ref{fig_gradient_intuition}, right subfigure):
\begin{equation}
\vspace{-3pt}
\label{retro_l1_grad}
    \begin{cases} 
        1 & for\  g_{\theta^T} > g_{\theta^{T_p}} \\
        2\kappa+1 & for\  g_{\theta^*}< g_{\theta^T} < g_{\theta^{T_p}}  \\
        -1 & for\ g_{\theta^T} < g_{\theta^*} 
    \end{cases}
\end{equation}
\noindent The choice of $L_1$-norm in the retrospective loss implies that the gradient is not defined at $g_{\theta^{T_p}}$ and $g_{\theta^*}$. However, it is evident from the gradient (slope) values that the minimum for $\mathcal{L}_{retrospective}$, when choosing the $L_1$-norm, is at $g_{\theta^*}$. The $L_1$-norm version of $\mathcal{L}_{retrospective}$ hence satisfies the consistency property.

It is not difficult to see that if one chooses the $L_2$-norm instead in the retrospective loss, Eqn \ref{retro_grad} becomes:
\begin{equation}
    2* \bigg((\kappa+1) *(g_{\theta^T} - g_{\theta^*}) -  
    \kappa*(g_{\theta^T} -  g_{\theta^{T_p}}) \bigg) 
\end{equation}
\noindent whose minimum lies at $g_{\theta^T} =  g_{\theta^*}+ \kappa *(g_{\theta^*}- g_{\theta^{T_p}})$. The $L_2$-norm version of $\mathcal{L}_{retrospective}$ hence does not satisfy the consistency property. We use the $L_1$-norm version of $\mathcal{L}_{retrospective}$ in all our experiments.

From another perspective, consider gradient descent, $\theta_{t+1}= \theta_{t} - \Delta \theta$ where the weight update is given for the task-specific loss by:
$\Delta \theta = \eta \nabla_{\theta} \mathcal{L}_{task} (g_{\theta})$. With the retrospective loss, the new weight update in a given iteration is given by:
\begin{equation}
\label{update_L_new_1}
\Delta \theta_{new}=\eta * \nabla (\mathcal{L}_{task}+ \mathcal{L}_{retrospective}) (g_{\theta})
\end{equation}
\begin{equation}
\label{update_L_new_2}
\implies \Delta \theta_{new}=\eta * \bigg(1 + \frac{\nabla\mathcal{L}_{retrospection}(g_{\theta})}{\nabla \mathcal{L}_{task}(g_{\theta})}\bigg)\nabla \mathcal{L}_{task}(g_{\theta})
\end{equation}
\begin{equation}
\label{update_L_new_3}
\implies \Delta \theta_{new}= \alpha * \Delta \theta
\end{equation}
where $\alpha= \bigg(1 + \frac{\nabla\mathcal{L}_{retrospection}(g_{\theta})}{\nabla \mathcal{L}_{task}(g_{\theta})}\bigg)$. Therefore, the new weight update could be thought of as having a variable learning rate $\eta * \alpha$.
It can be seen from Eqns \ref{retro_l1_grad} and \ref{update_L_new_2}, that an additional slope of $\frac{2\kappa+1}{\nabla \mathcal{L}_{task}}$ has been added to the loss surface for $ g_{\theta^T} < g_{\theta^{T_p}}$ and $g_{\theta^T} > g_{\theta^*}$. This increases the learning rate modifier $\alpha$ in this interval. It could also be seen that when $g_{\theta^T} > g_{\theta^*}$, $\alpha$ changes by $\frac{-1}{\nabla \mathcal{L}_{task}}$ instead of $\frac{2\kappa+1}{\nabla \mathcal{L}_{task}}$. This asymmetric slope \cite{NIPS2019_8524} change on the two sides leads to widening of the global minimum valley and reduces the oscillatory behavior in spite of increase in effective learning rate. In case $g_{\theta^T}$ is stuck at $g_{\theta^T} < g_{\theta^*}$, then after a duration of $T_{lag}$, $ g_{\theta^{T_p}}$ is updated and this changes the learning rate by $\frac{2\kappa+1}{\nabla \mathcal{L}_{task}}$ (instead of $\frac{-1}{\nabla \mathcal{L}_{task}}$) for this interval and widens the valley for $g_{\theta^T} > g_{\theta^*}$. This helps in taming the oscillatory behavior. We provide empirical evidence (in our results in Sec \ref{sec_expts}) that the proposed retrospective loss also takes models to a better minimum, supporting this analysis.\\
\vspace{-6pt}


\noindent \textbf{Update Frequency.} In practice, while implementing the retrospective loss, we define a retrospective update frequency, $F$, which gives an upper bound difference of the previous training step ($T_p$) from the current training step T at which we compute the retrospective loss. 
We use $T_p = F*\floor{T/F}$ as the time step for retrospection in this work, and show gains in efficiency of training. One could however mine for $T_p$ intelligently to further improve the performance, which we leave for future work.\\
\vspace{-6pt}

\noindent \textbf{Connection with Triplet Loss.}
The triplet loss (\citep{chechik2010large, schroff2015facenet, hoffer2015deep}) has been proposed and used extensively over the last few years to learn high-quality data embeddings, by considering a triplet of data points, $\textbf{x}_a$ (anchor point), $\textbf{x}_p$ (point from the positive/same class as the sample under consideration), and $\textbf{x}_n$ (point from the negative class/class different from the sample under consideration). The loss is then defined as:
\vspace{-3pt}
\begin{equation}
    \label{eqn_triplet_loss}
    \max \left( {\Vert g_a- g_p \Vert}^2 - {\Vert g_a - g_n \Vert}^2 + m, 0\right)\ 
\end{equation}
\noindent where $g$ is the neural network model, and $m$ is a minimum desired margin of separation. The triplet loss, inspired by contrastive loss  \citep{hadsell2006dimensionality}, attempts to learn parameters $\theta$ of a neural network in such a way that data points belonging to the same class are pulled together closer than a data point from another class. One could view the proposed retrospective loss as a triplet loss in the parameter space. While the traditional triplet loss consider a triplet of data samples, we consider a triplet of parameters, $\theta^T$, $\theta^*$, and $\theta^{T_p}$, where $\theta^{T_p}$ is obtained from previous parameter states in time.\\
\vspace{-6pt}

\noindent \textbf{Connection with Momentum.}
Viewing retrospective loss from the perspective of previous gradients in the training trajectory, one can connect it to the use of momentum, although more in a contrasting sense. The use of momentum and variants such as Nesterov momentum in training neural networks use the past gradient, say at $\theta^{T-1}$ or the gradient over the previous few steps, at $\{\theta^{T-q},\cdots,\theta^{T-1}\},q>0$), while updating the parameters in the current step. This assumes local consistency of the direction of the gradient update in the training trajectory, and that one can use these previous directions to get a more robust estimate of the gradient step to be taken currently. In contrast, retrospective loss leverages this idea from the opposite perspective, viz., the direction of the gradient update is \textit{only} locally similar, and hence the parameter state, $\theta^{T_p}$ farther away from the current state $\theta^{T}$, is an indicator of what the next parameter state must be far from. Physically speaking, this raises interesting discussions, and the possibility of analyzing retrospective loss as a thrust obtained from an undesirable parameter state, as opposed to momentum, which we leave as directions of future work and analysis at this time.

\begin{figure*}[t]
\begin{center}
\minipage{0.25\textwidth}
\includegraphics[scale=0.13]{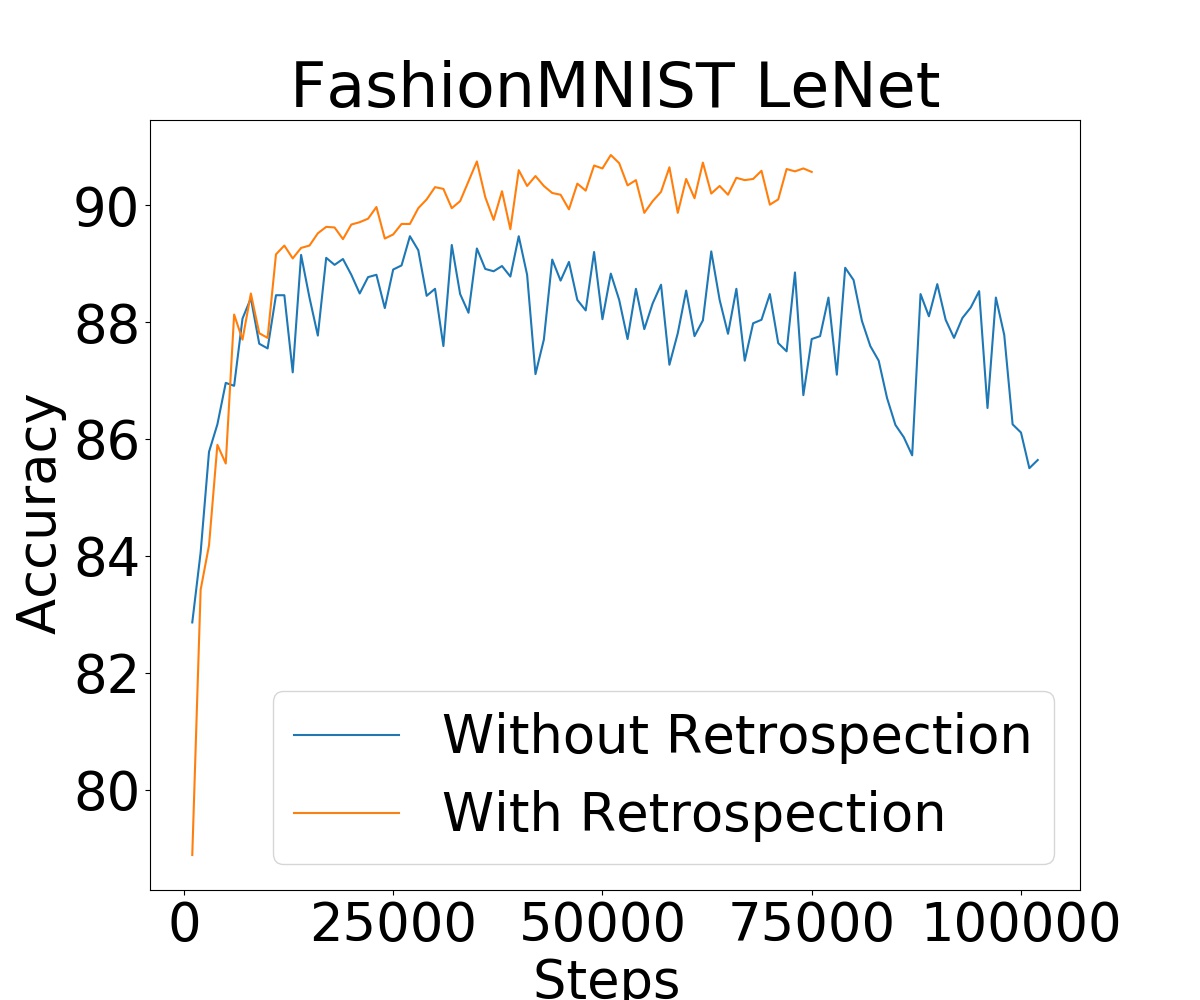}
\endminipage\hfill
\minipage{0.25\textwidth}
\includegraphics[scale=0.13]{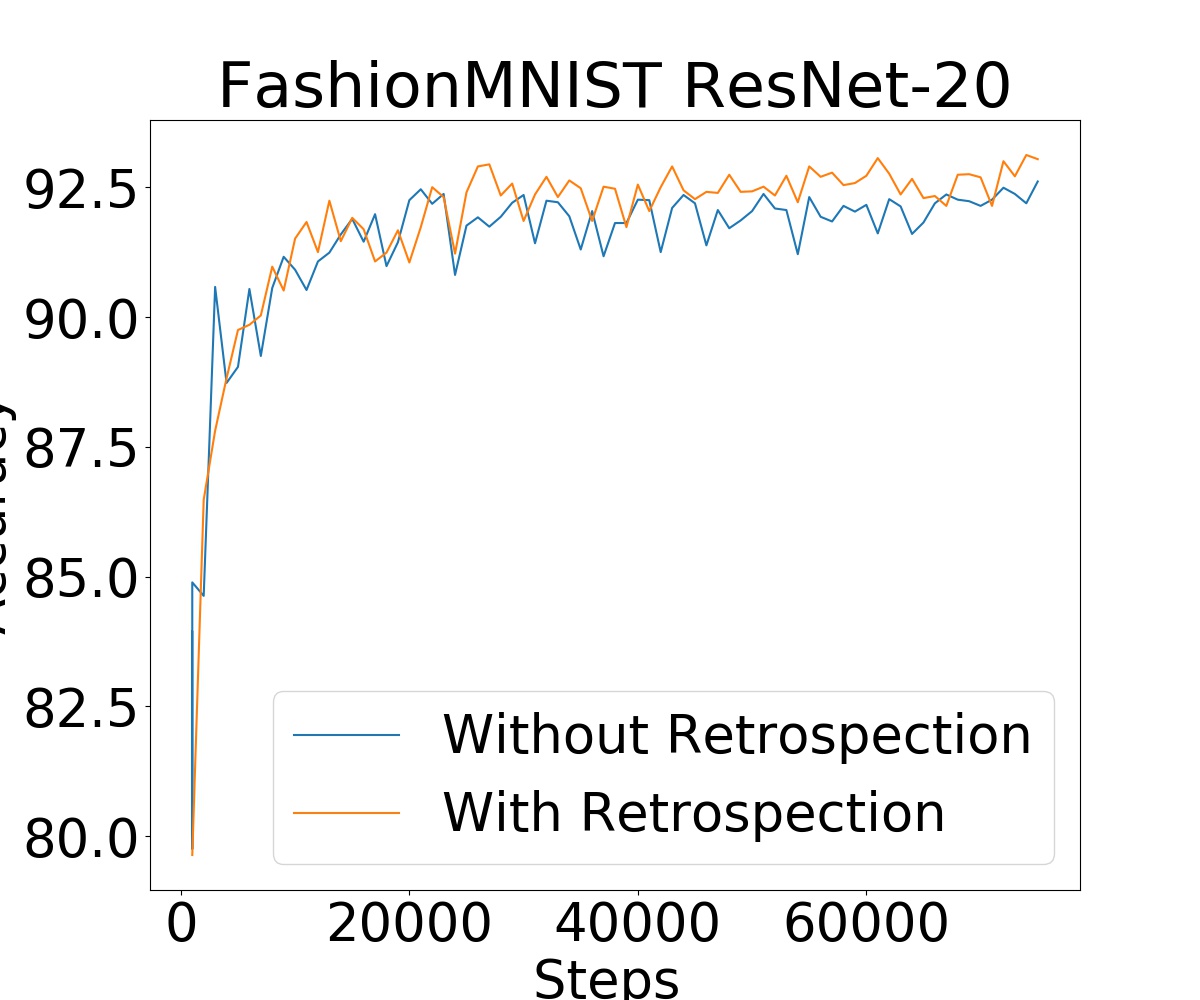}
\endminipage\hfill
\minipage{0.25\textwidth}
\includegraphics[scale=0.13]{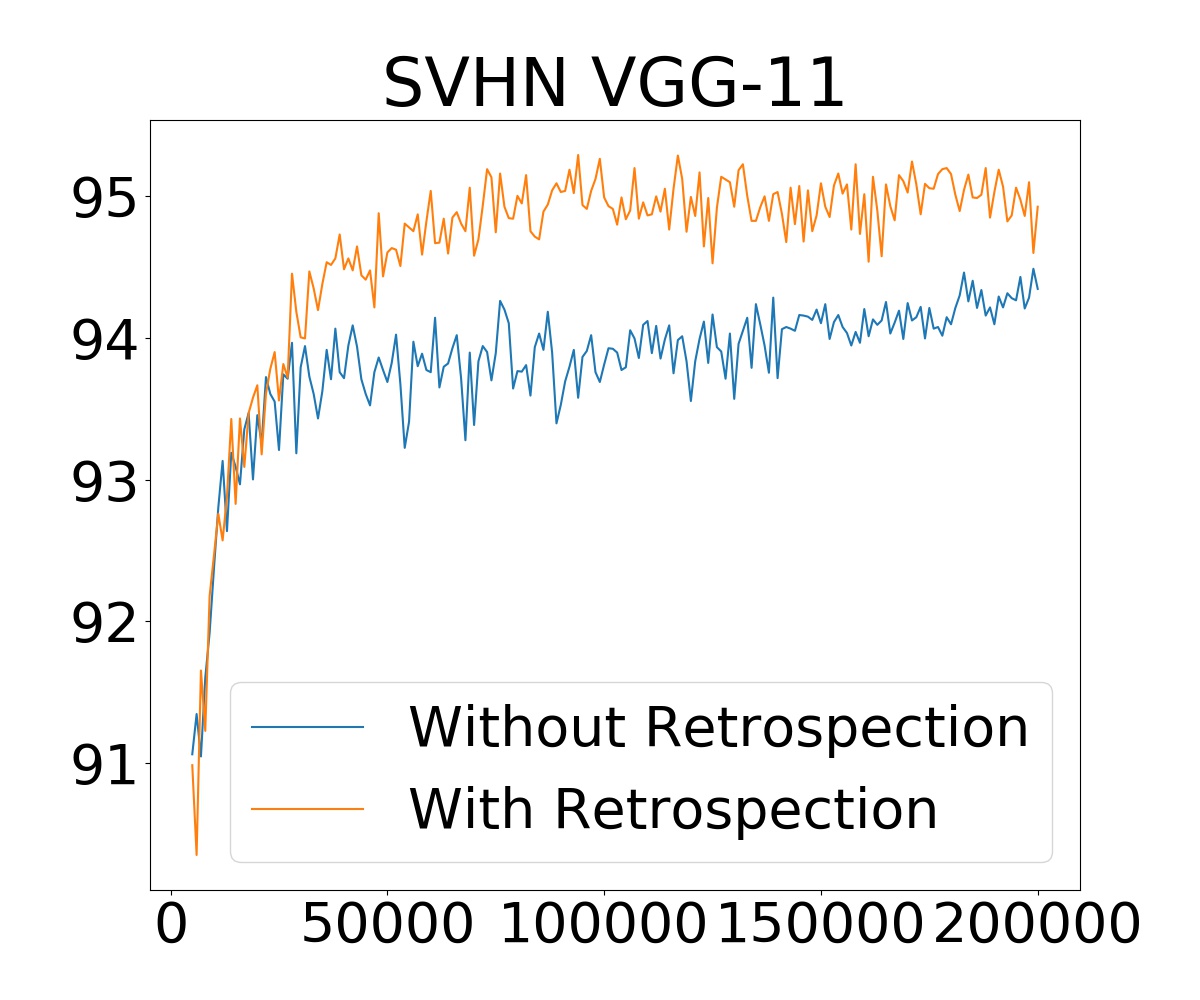}
\endminipage\hfill
\minipage{0.25\textwidth}
\includegraphics[scale=0.13]{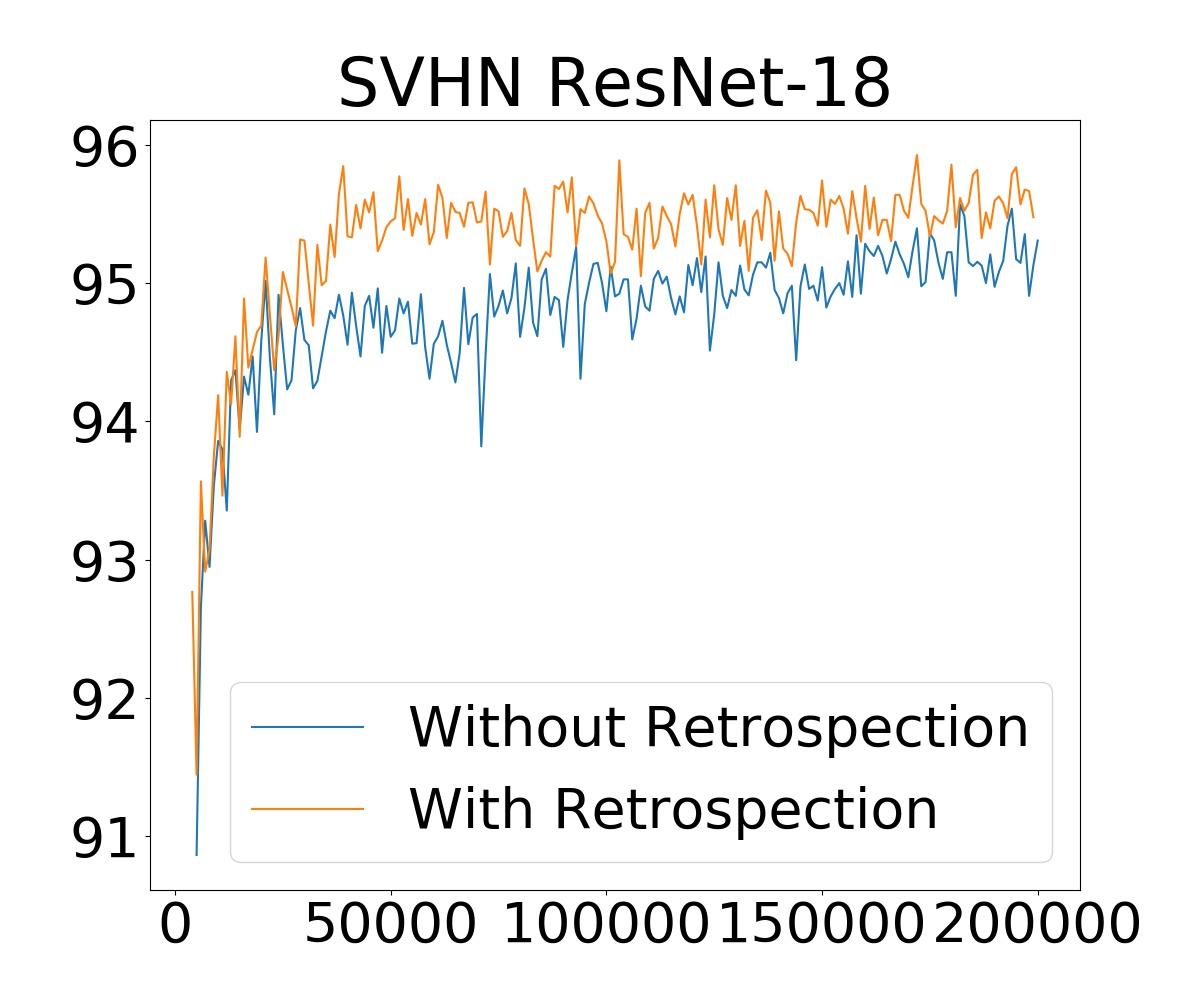}
\endminipage
\vspace{-4pt}
\caption{Image classification: Evolution of test accuracy with and without retrospective loss on F-MNIST and SVHN datasets}
\vspace{-8pt}
\label{classify-figure}
\end{center}
\end{figure*}

\section{Experiments and Results}
\label{sec_expts}
We study the usefulness of the proposed retrospective loss by conducting experiments on a wide range of data domains, including images (Sec \ref{subsec_image_expts}), text (Sec \ref{subsec_text_expts}), speech (Sec \ref{subsec_speech_expts}), and graphs (Sec \ref{subsec_graph_expts}) = to show its usefulness on DNN models across application domains. We also consider multiple image tasks, including image classification, few-shot classification and image generation in Sec \ref{subsec_image_expts}. 
In all our experiments, we ensured that the DNN was trained with and without retrospective loss using the same weight initialization, for fairness of comparison. As stated (and substantiated) in Sec \ref{sec_methodology}, we use the $L_1$-norm as the choice of norm for retrospective loss in all our implementations. The hyperparameter $\kappa$ is set to 2. In Sec \ref{sec_analysis}, we further study the impact of hyperparameter choices (update frequency, choice of norm, warm-up period, batch size) in the experiments, as well as comparison with methods such as momentum. When retrospection is used without warm-up, the guidance parameters, $\theta^{T_p}$, are initialized at random. All the experiments were performed using the \textbf{Adobe Sensei} platform.

\subsection{Experiments on Image Data}
\label{subsec_image_expts}
\subsubsection{\textbf{Image Classification}}
We carry out image classification experiments on multiple benchmark datasets including Fashion-MNIST \citep{fmnist}, SVHN \citep{svhn}, CIFAR-10 \citep{cifar-10} and TinyImageNet \citep{tiny_imagenet} datasets. The retrospective loss here uses activations of the softmax layer for $g_{\theta}(\textbf{x}_i)$. For each dataset, we use widely used architectures like ResNets and VGG to show the usefulness of the proposed loss. We now describe the experimental setup for each dataset, and the results are compiled in Table \ref{img-classify-table}.\\
\vspace{-7pt}

\noindent \textit{Fashion-MNIST.} 
For experiments on FMNIST, we use LeNet \citep{leNet} and ResNet-20 \citep{resnet} architectures.  Models in each experiment are trained to convergence using the SGD optimizer (lr=0.1, momentum=0.5, mini-batch=32) running over 70,000 steps. Results in Figure \ref{classify-figure} (a)-(b) show that using the retrospective loss results in better and faster convergence (with significant gains especially on LeNet). The t-SNE plots obtained using the model trained using retrospective loss also show improved discriminability, as shown in Fig \ref{fig_tsne_plots_lenet}.\\
\vspace{-6pt}

\begin{table}[h]
\centering
\begin{tabular}{c|c|c|c}
\hline
Dataset                   & Model      & Original    & Retrospective \\
\hline 
\multirow{2}{*}{F-MNIST}  & LeNet      & 10.8        & 9.4           \\
                          & ResNet-20  & 7.6         & 6.8           \\
\hline
\multirow{2}{*}{SVHN}  & VGG-11     & 5.54        & 4.70          \\
                          & ResNet-18  & 4.42        & 4.06          \\
\hline
\multirow{3}{*}{CIFAR-10} & ResNet-44  & 6.98 (7.17) & 6.55          \\
                          & ResNet-56  & 6.86 (6.97) & 6.52         \\
                          & ResNet-110 & 6.55 (6.61) & 6.29\\
\hline
\multirow{2}{*}{Tiny-ImageNet} & ResNet-18  & 45.17 & 43.14  \\
& ResNet-50  & 35.13  & 34.69 \\
\hline
\end{tabular}
\caption{Image classification results: Test error using retrospective loss on F-MNIST, SVHN,  CIFAR-10 and Tiny-ImageNet datasets.}
\vspace{-10pt}
\label{img-classify-table}
\end{table}

\noindent \textit{SVHN.} For experiments on SVHN, we use VGG-11 \citep{vgg} and ResNet-18 \citep{resnet} architectures. Models in each experiment are trained to convergence using the SGD optimizer (lr=0.001, momentum=0.9, mini-batch=100) running over 200,000 steps. Results in Figure \ref{classify-figure} (c)-(d) show that using the retrospective loss results in significant improvement in performance almost all through training. No warm-up period was used in these experiments for FMNIST and SVHN datasets, with a retrospective update frequency of fifty steps.\\
\vspace{-7pt}

\noindent \textit{CIFAR-10.} For experiments on CIFAR-10 \citep{cifar-10}, we use larger variants of ResNet including ResNet - 44, 56, 110, following \citep{resnet}. Models in each experiment are trained for 200 epochs, using the training configuration (mini-batch, lr policy) detailed in \citep{resnet}. Here, we observe that using the retrospective loss in later stages of training results in best improvement in performance. Correspondingly, the retrospective loss is introduced after a warm-up of 150 epochs and the retrospective update frequency there on is one epoch. 
In Table \ref{img-classify-table}, we also mention (in parantheses) the error rates for the corresponding experiments reported in the original work \citep{resnet}.\\
\vspace{-7pt}

\noindent \textit{Tiny ImageNet.} For experiments on TinyImageNet \citep{tiny_imagenet}, we use ResNet - 18, 56. Models in each experiment are trained using the training configuration (mini-batch, lr policy) detailed in \citep{resnet_tinyimg}. The retrospective update frequency in one epoch. The quantitative results reported in Table \ref{img-classify-table} show consistent improvement in performance across datasets and architectures when using the retrospective loss. In general, we obtained these results with minimal tuning of hyperparameter choices in the retrospective loss.


\subsubsection{\textbf{Few-shot Classification}}
\label{subsec_few_shot}
We next conducted experiments on the task of few-shot image classification using the widely used CUB-200 \citep{cubdata} benchmark dataset. CUB-200 consists of 11,788 images from 200 bird species. In few-shot learning, the ability of a model is measured by its performance on $n$-shot, $k$-way tasks where the model is given a query sample belonging to a new, previously unseen class and a support set, $S$, consisting of $n$ examples each from $k$ different unseen classes. The model then has to determine which of the support set classes the query sample belongs to. We use the 5-way 5-shot setting in our experiments and compare against CloserLook \citep{closerlookfewshot}, a recent state-of-the-art work, and ProtoNet \citep{protonet}, another popular work from the domain. Our experimental setup follows \citep{closerlookfewshot}, and implementations use publicly available code on \citep{fewshotgithub}. Our experiments include backbones of Conv4, Conv6 and ResNet34, as in \citep{closerlookfewshot}. For our experiments, each model (with and without retrospective loss) is trained on ProtoNet \citep{protonet} for 400 epochs and on CloserLook \citep{closerlookfewshot} for 200 epochs.For Conv4 and Conv6 configurations, retrospection is introduced without any warm-up period (zero epochs). For ResNet34, a warm-up period of 280 epochs for ProtoNet and 150 epochs for CloserLook is used. For all experiments, the retrospective update frequency is one epoch. For CloserLook, we report comparative performance with Baseline++, their best performing variant. The results, reported in Table \ref{fewshot-table}, highlight that training with retrospective loss results in improved classification accuracy for all backbone configurations on both CloserLook and ProtoNet. \footnote{Results in some experiments on the original configuration do not match values (are higher or lower) reported in \citep{closerlookfewshot} even after using official code and same training config. However, we ensure consistency of comparison by using the same initializations for original and retrospective settings.} We note in particular that the results obtained herein for few-shot classification with retrospective loss outperform the state-of-the-art results in \citep{closerlookfewshot}.

\begin{table}
\small
\centering
\begin{tabular}{c|cc|cc}
\hline
Model    & \multicolumn{2}{|c}{protonet}                  & \multicolumn{2}{|c}{closerlook}             \\
\hline
         & Original                           & Retrospective            & Original                        & Retrospective            \\
\hline 
Conv4    & 75.26 $\pm$ 1.05  & 78.64 $\pm$ 1.25 &                   79.03 $\pm$ 0.63         &  79.95 $\pm$ 0.75           \\
Conv6    & 80.71 $\pm$ 1.55 & 81.78 $\pm$ 1.40 &            81.05 $\pm$ 0.55      &   81.35 $\pm$ 0.30   \\
ResNet34 & 88.75 $\pm$ 1.01 & 89.99 $\pm$ 1.13  & 82.23 $\pm$ 0.59 & 83.11 $\pm$ 0.55 \\
\hline
\end{tabular}
\caption{Few-shot classification: Test accuracy with and without retrospective loss on CUB dataset. We report mean and std deviation over 10 runs of random initializations.}
\vspace{-9pt}
\label{fewshot-table}
\end{table}

\begin{figure}[h]
\begin{center}

\includegraphics[scale=0.16]{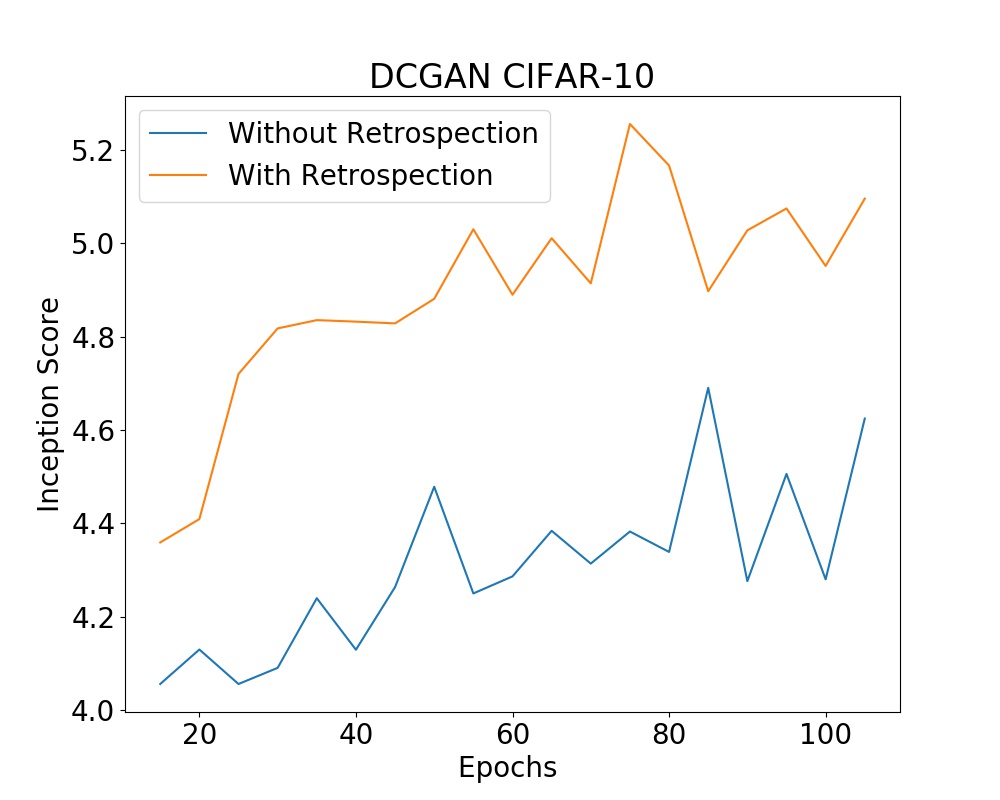}
\includegraphics[scale=0.16]{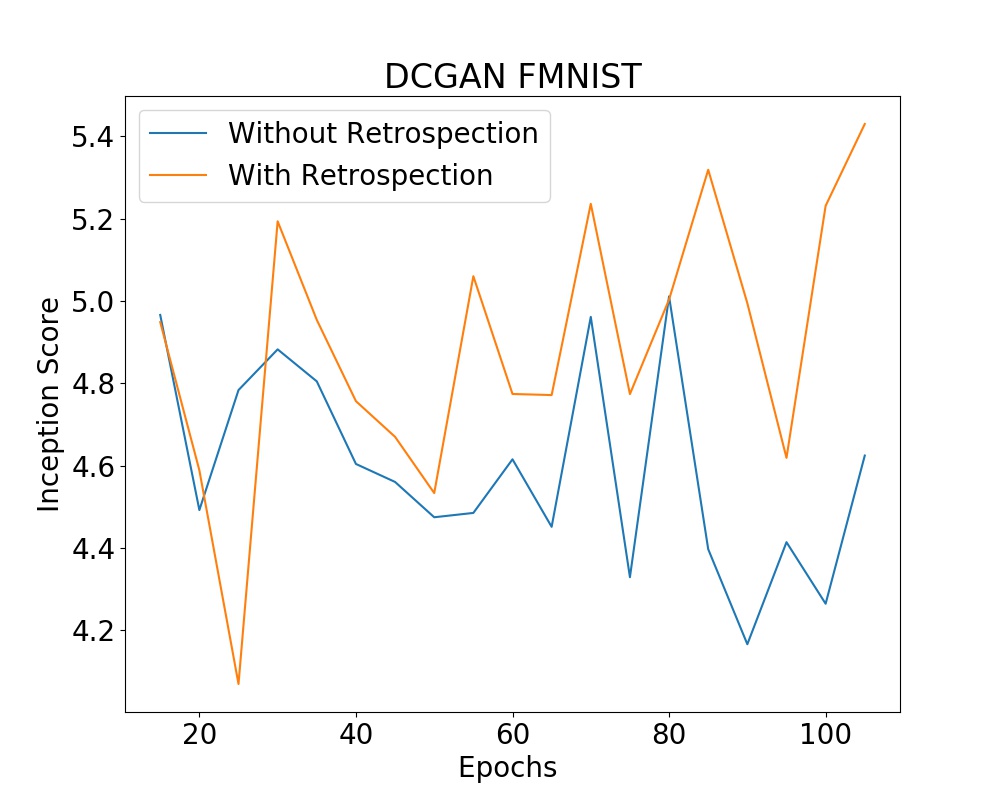}\\

\includegraphics[scale=0.16]{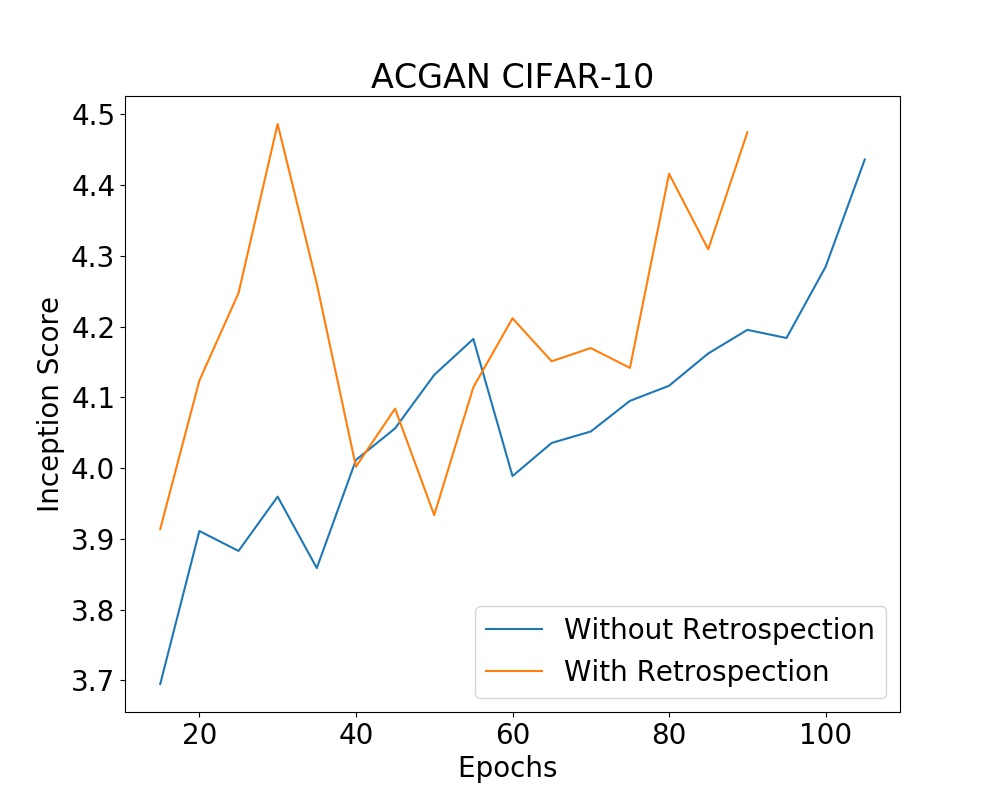}
\includegraphics[scale=0.16]{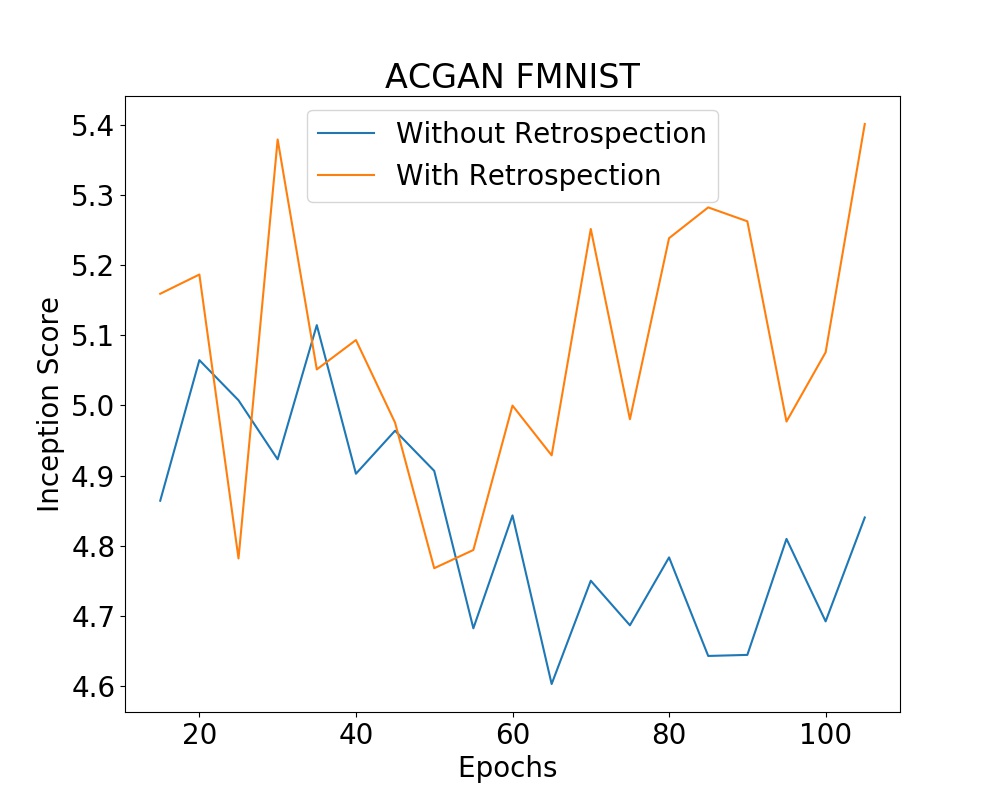}\\

\includegraphics[scale=0.16]{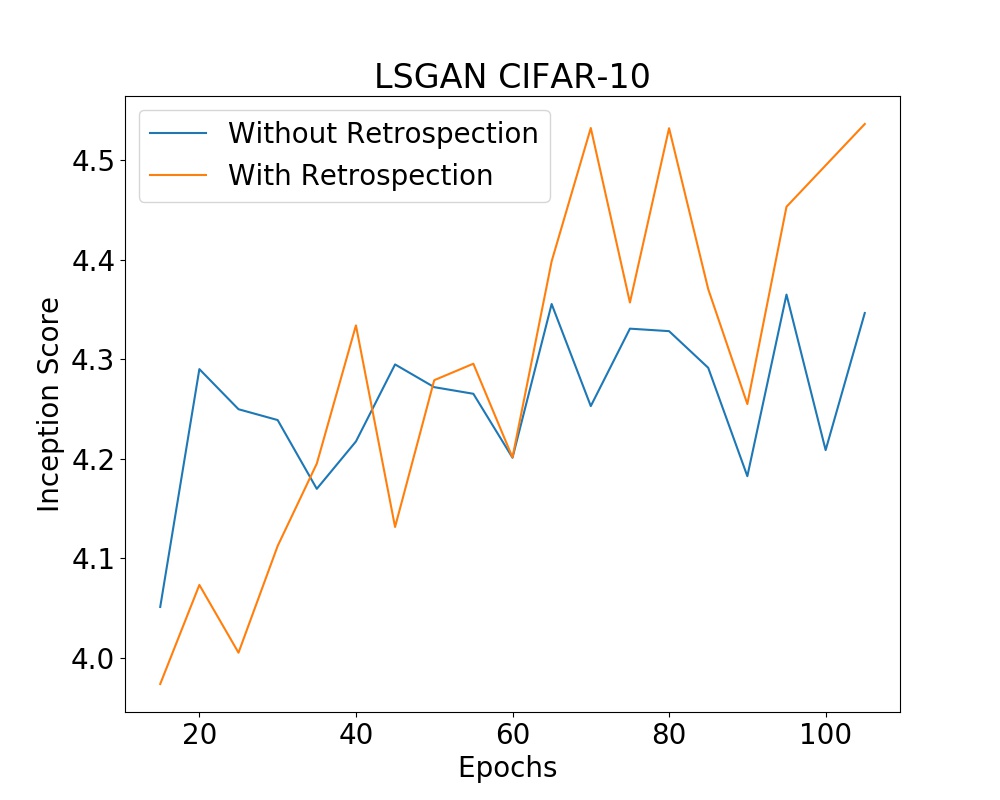}
\includegraphics[scale=0.16]{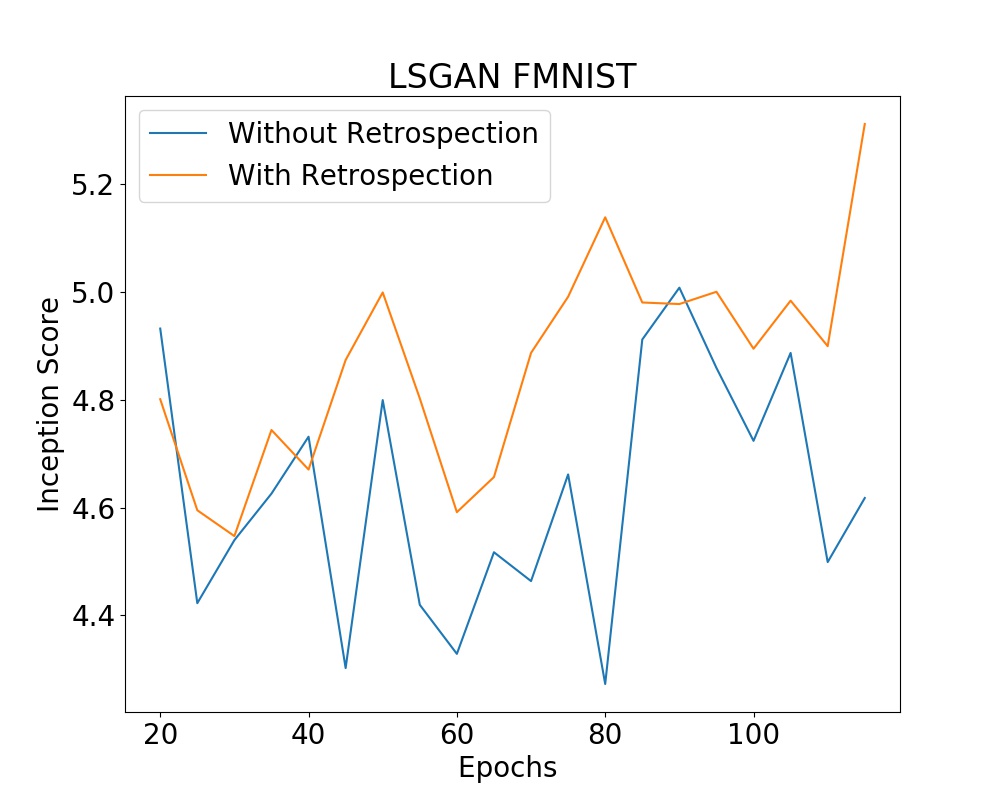}

\end{center}
\vspace{-6pt}
\caption{Image generation task: Evolution of Inception Scores using retrospection on CIFAR-10 \citep{cifar-10}(Col 1) and FMNIST \citep{fmnist} (Col 2) datasets using DCGAN \citep{dcgan} (Row 1), ACGAN \citep{acgan} (Row 2), LSGAN \citep{lsgan} (Row 3).}
\vspace{-10pt}
\label{fig_gan_inception}
\end{figure}
\subsubsection{\textbf{Image Generation}}
\label{subsec_gan}
\begin{figure}[h]
    \centering
    \includegraphics[scale=0.33]{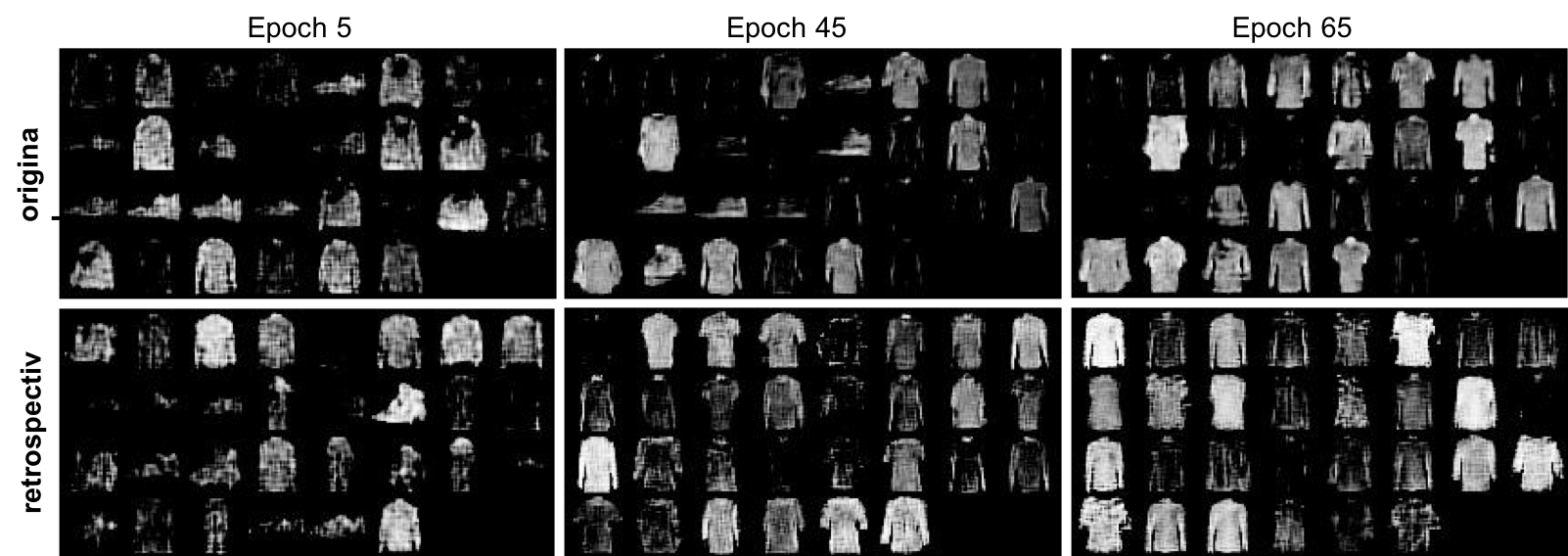}
    \vspace{-6pt}
    \caption{Images generated over training epochs when ACGAN \citep{acgan} is trained on FMNIST dataset: (a) without retrospection (Row 1) (b) with retrospection (Row 2). Note the significant increase in quality with retrospective loss (best visible when zoomed)}
    \vspace{-9pt}
    \label{fig:acgan-results}
\end{figure}

We also performed experiment on the image generation task with Generative Adversarial Networks (GANs) on  FMNIST \citep{fmnist} and CIFAR-10 \citep{cifar-10} datasets. Our study considers both unconditional (DCGAN, LSGAN) and conditional (ACGAN) variants of GANs. 
We adapt implementations from \citep{dcganpytorch} for LSGAN \citep{lsgan} and DCGAN \citep{dcgan}, and from \citep{acgan_pytorch} for ACGAN \citep{acgan}. We train the generator and discriminator for 100 epochs, with initial learning rate of 0.0002 on mini-batches of size 64 using Adam optimizer. We report performance using Inception Score \citep{inception-score}, a standard metric for evaluating GANs. (The Inception score is computed using the implementation in \citep{inception-score-pytorch} with predictions for CIFAR-10 generated using network in \citep{inceptionnet} and features for FMNIST using network in \citep{alexnet}).
For all experiments, the retrospective loss is initialized without any warm-up period (zero epochs). The loss is computed on outputs of the discriminator and is used to train the generator model. 
The retrospective update is used six times in one epoch. The scaling parameter, $\kappa$ is set to 4. For ACGAN \citep{acgan}, which is conditional, the retrospective loss consists of both adversarial loss and class loss components. 
Figure \ref{fig_gan_inception} presents the Inception score plots, and shows consistent improvement of Inception scores when training with retrospective loss. Figure \ref{fig:acgan-results} presents images generated over epochs when training ACGAN \citep{acgan}, with and without retrospection, on F-MNIST \citep{fmnist}, which again supports the use of retrospective loss.

\subsection{Experiments on Text Data}
\label{subsec_text_expts}
We perform text classification experiments on the task of emotion detection in dyadic conversations using DialogueRNN \citep{dialogueRNN}, a recent state-of-the-art work, which is composed of an attentive network consisting of three Gated Recurrent Units (GRUs). Our experiments are conducted on the AVEC \citep{avec} and IEMOCAP \citep{iemocap} datasets. While the datasets are multi-modal (image and text), we follow the DialogueRNN work \citep{dialogueRNN} and restrict our inputs to text alone. The text data is pre-processed to obtain $n$-gram features as in \citep{dialogueRNN}. We follow the same train-test split and training configurations in \citep{dialogueRNN} too. Performance comparison is reported on $BiDialogueRNN_+Att$, the best performing variant from the original work. For experiments on IEMOCAP, models in each experiment are trained for 60 epochs on cross-entropy objective with F1-Score and accuracy as performance metrics. No warm-up was found to be required on this dataset when using retrospective loss. On AVEC, models in each experiment are trained for 100 epochs using MSE loss with MSE and pear-score(r) as the performance metrics. Here, introducing retrospective loss after a warm-up of 75 epochs provides best performance. For experiments on both IEMOCAP and AVEC, the retrospective update frequency is one epoch. Experiments are conducted using the official code repository for this work \citep{DialogueGithub}. Results in Table \ref{emotion-table} show that using the retrospective loss when training DialogueRNN improves performance significantly on both IECOMAP and AVEC datasets.
\begin{table}[h]
\footnotesize
\begin{tabular}{c|c|c|c|c}
\hline
\multirow{2}{*}{Method} & \multicolumn{2}{|c|}{IECOMAP} & \multicolumn{2}{|c}{AVEC}          \\
                        & F1-Score     & Accuracy     & MSE              & R (Pear Score) \\
\hline
Original                & 62.60 $\pm$ 0.9 & 62.70 $\pm$ 0.7 & 0.1798 $\pm$ 0.0005 & 0.317 $\pm$ 0.007 \\
Retrospective           & 64.40 $\pm$ 0.4 & 64.97 $\pm$ 0.5 & 0.1772 $\pm$ 0.0006 & 0.332 $\pm$ 0.008 \\
\hline
\end{tabular}
\caption{Text classification task: Performance on using retrospective loss for dyadic emotion recognition with DialogueRNN. We report mean and std deviation over 10 runs of random initializations.}
\vspace{-20pt}
\label{emotion-table}
\end{table}




\subsection{Experiments on Speech Data}
\label{subsec_speech_expts}
We perform speech recognition experiments on the Google Commands \citep{gcommands} dataset, which consists of 65,000 utterances, where each utterance is about one-second long and belongs to one of 30 classes. The classes correspond to voice commands such as yes, no, down, left, as pronounced by a few thousand different speakers. We follow \citep{zhang2017mixup} to pre-process the utterances, where we first extract normalized spectrograms from the original waveforms at a sampling rate of 16 kHz and subsequently zero-pad the spectrograms to normalize their sizes at $160 \times 101$. For this study, we use the LeNet\citep{leNet} and VGG-11\citep{vgg} architectures on the obtained spectrograms. We train each model for 30 epochs with mini-batches of 100 examples, using Adam as the optimizer. Training starts with a learning rate of $3$x$10^{-3}$ and is divided by 10 every 10 epochs. The retrospective loss is introduced after a warm-up period of 8 epochs. The retrospection update frequency is half an epoch. The results are reported in Table \ref{speech-table}, and show that training using retrospective loss decreases error rate for both LeNet \citep{leNet} and VGG-11 \citep{vgg} on both validation and testing sets on this speech task too.
\begin{table}
\small
\centering
\begin{tabular}{c|c|c|c|c}
\hline
Model & \multicolumn{2}{|c|}{Validation Set} & \multicolumn{2}{|c}{Testing Set} \\
\hline
                       & Original               & Retrospective          & Original              & Retrospective         \\
\hline
LeNet                  & 9.77 $\pm$ 0.05           & 9.60 $\pm$ 0.03            & 10.26 $\pm$ 0.05         & 9.86 $\pm$ 0.04          \\
\hline
VGG-11                 & 5.15 $\pm$ 0.08           & 4.37 $\pm$ 0.04            & 5.03 $\pm$ 0.06          & 4.16 $\pm$ 0.05  \\
\hline
\end{tabular}
\caption{Speech recognition task: Classification error using retrospective loss on the Google Commands dataset. We report mean and std deviation over 10 runs of random initializations.} 
\vspace{-10pt}
\label{speech-table}
\end{table}



\subsection{Experiments on Graph Data}
\label{subsec_graph_expts}
We study the impact of using retrospective loss on the task of semi-supervised node classification on the popular \textit{CORA} and \textit{CITESEER} datasets \citep{graph-datasets}. For our experiments, we use two different models: ARMA \citep{arma} (a recent state-of-the-art method) and GCN \citep{gcn}, another well-known method for graph analysis in recent times. Our implementations follow \citep{torchgeometric} for this study. Performance is reported by averaging results over 30 experimental runs, each of which involves training the model for 100 epochs. No warm-up period was found to be required for these experiments. The hyperparameters, $F$ and $\kappa$, used for training on both CORA and CITESEER are: (a) GCN: $F$ = 2, $\kappa$ = 4; (b) ARMA: $F$ = 1, $\kappa$=3. Table \ref{graph-data} presents the quantitative results of using retrospective loss, which corroborates our claim of the usefulness of the loss across domains.
\begin{table}[h]
\small
\begin{center}
\begin{tabular}{c|cc|c|cc}
\hline
Dataset                   & \multicolumn{2}{|c|}{Config}              & GCN        & \multicolumn{2}{|c}{ARMA}       \\
\hline
\multirow{2}{*}{CORA}     & \multicolumn{2}{|c|}{Original}      & 80.85 $\pm$ 0.53 & \multicolumn{2}{|c}{78.53 $\pm$ 1.5}  \\
                          & \multicolumn{2}{|c|}{Retrospective} & 81.23 $\pm$ 0.27 & \multicolumn{2}{|c}{79.45 $\pm$ 1.15} \\
\hline
\multirow{2}{*}{CITESEER} & \multicolumn{2}{|c|}{Original}      & 70.65 $\pm$ 0.93 & \multicolumn{2}{|c}{63.63 $\pm$ 1.3}  \\
                          & \multicolumn{2}{|c|}{Retrospective} & 71.25 $\pm$ 0.75 & \multicolumn{2}{|c}{64.22 $\pm$ 1.2}  \\
\hline
\end{tabular}
\end{center}
\caption{Graph node classification: Performance using retrospective loss on CORA and CITESEER graph datasets. We report mean and std deviation over 10 runs of random initializations.}
\vspace{-20pt}
\label{graph-data}
\end{table}
\section{Analysis}
\label{sec_analysis}
In this section, we present ablation studies to analyze the impact of different hyperparameters - batch size, optimizer, retrospective update frequency ($F$), warmup period and norm. The studies are conducted on the image classification task on the F-MNIST \citep{fmnist} dataset using LeNet \citep{leNet} architecture. The default training configurations are used from Sec \ref{subsec_image_expts} In all these studies, DNNs are initialized with the same weights to ensure fairness of comparison.\\
\vspace{-6pt}

\begin{figure}[h]
  \begin{minipage}[b]{0.2\textwidth}
    \hspace{-2pt}
    \includegraphics[scale=0.11]{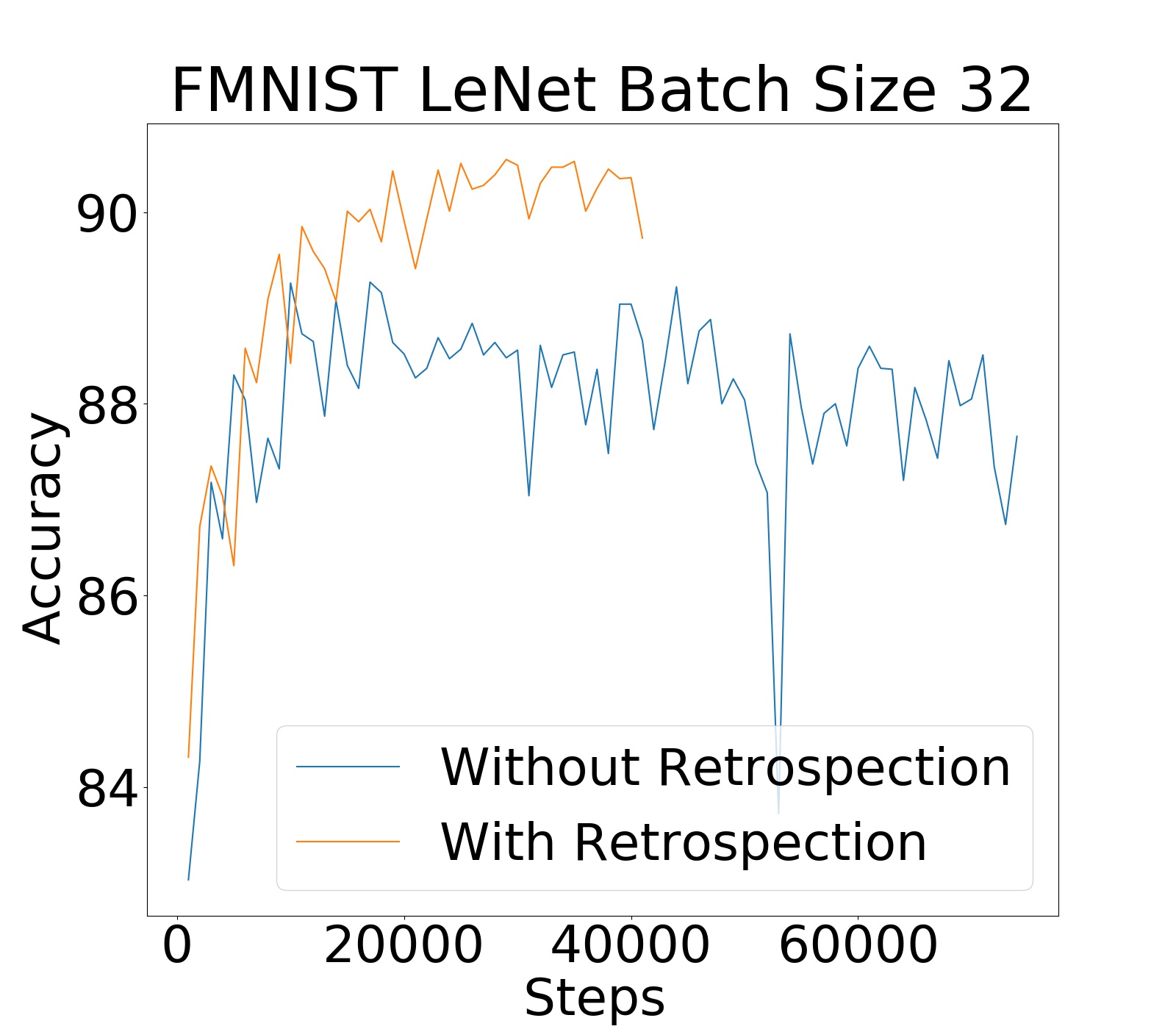}
  \end{minipage}
  \hspace{0.4cm}
  \begin{minipage}[b]{0.2\textwidth}
    \includegraphics[scale=0.11]{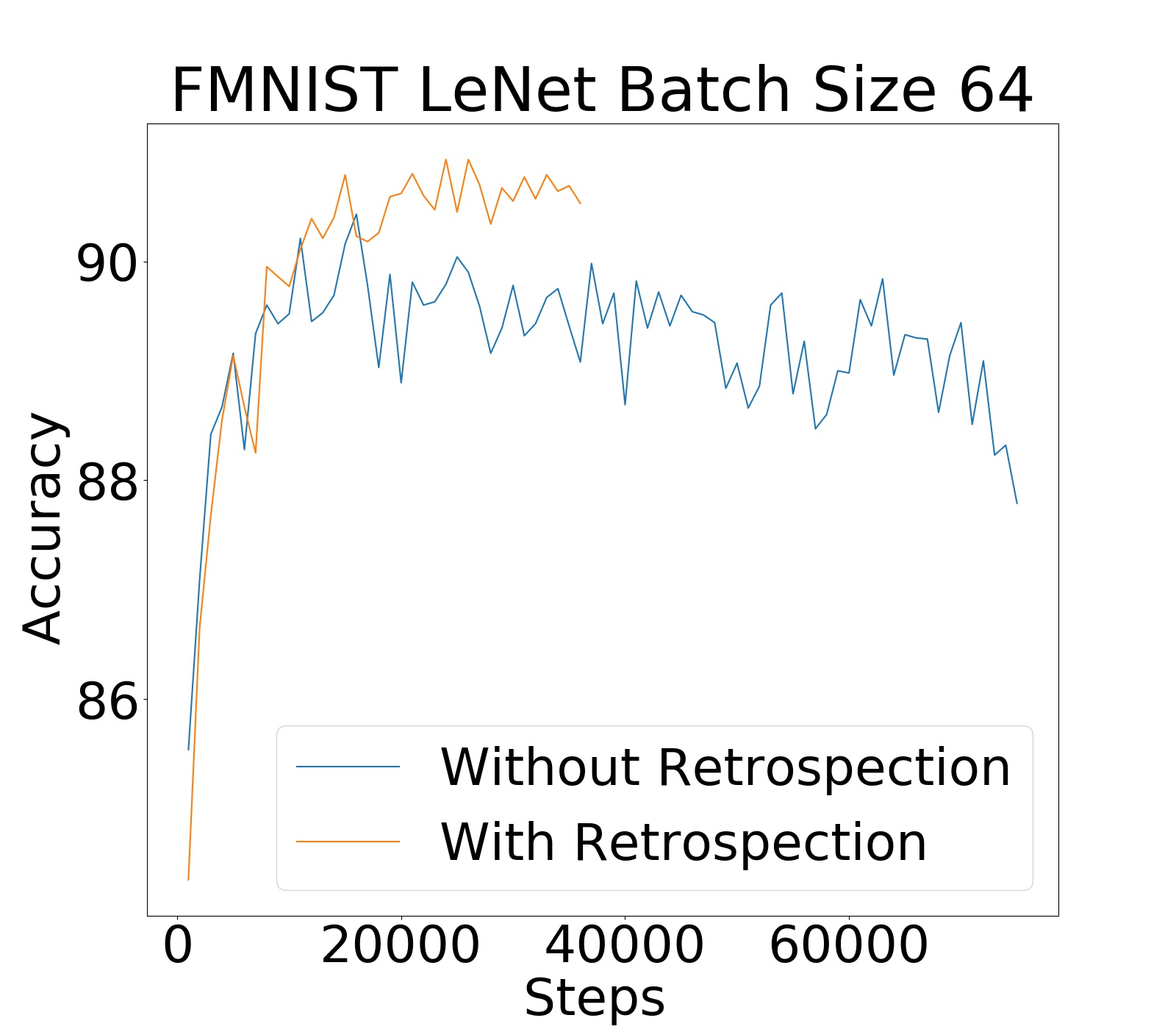}
  \end{minipage}
  
    \begin{minipage}[b]{0.2\textwidth}
    \includegraphics[scale=0.11]{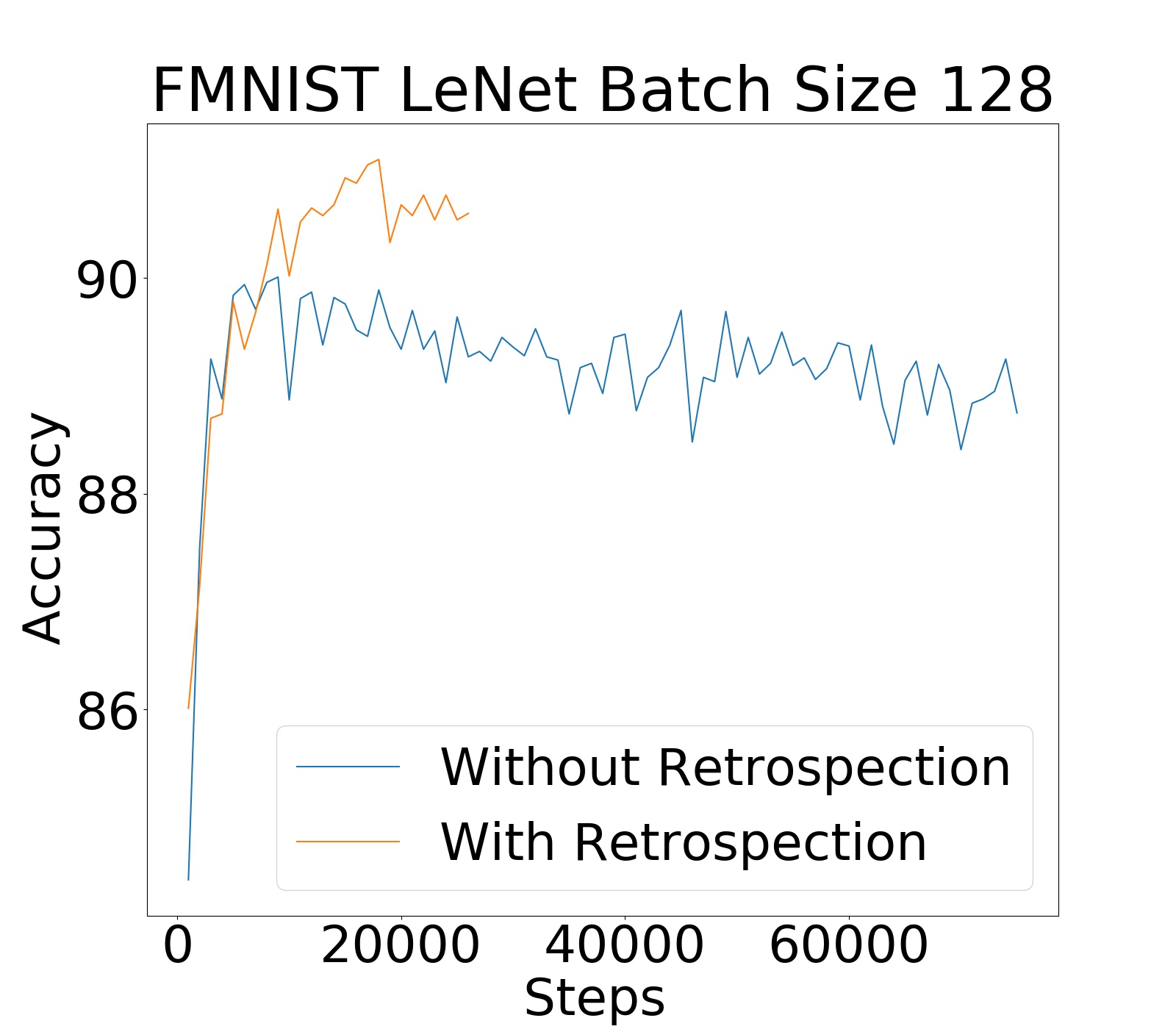}
  \end{minipage}
  \vspace{-5pt}
\caption{Classification performance using retrospection on LeNet\citep{leNet} across different batch sizes on FMNIST \citep{fmnist}}
\vspace{-9pt}
\label{fig:ablation-batchsize}
\end{figure}

\noindent \textbf{Choice of Batch Size.} We perform experiments to analyze the impact of the choice of mini-batch size when using the retrospective loss. We consider batch sizes - 32, 64, 128 in this study. The results are presented in Figure \ref{fig:ablation-batchsize}, which shows that we obtained improved performance using  retrospective loss across the considered batch sizes, leading us to infer that the loss is not sensitive to batch sizes.

\begin{figure}[h]
\hspace{-2pt}
  \begin{minipage}[b]{0.2\textwidth}

    \includegraphics[scale=0.11]{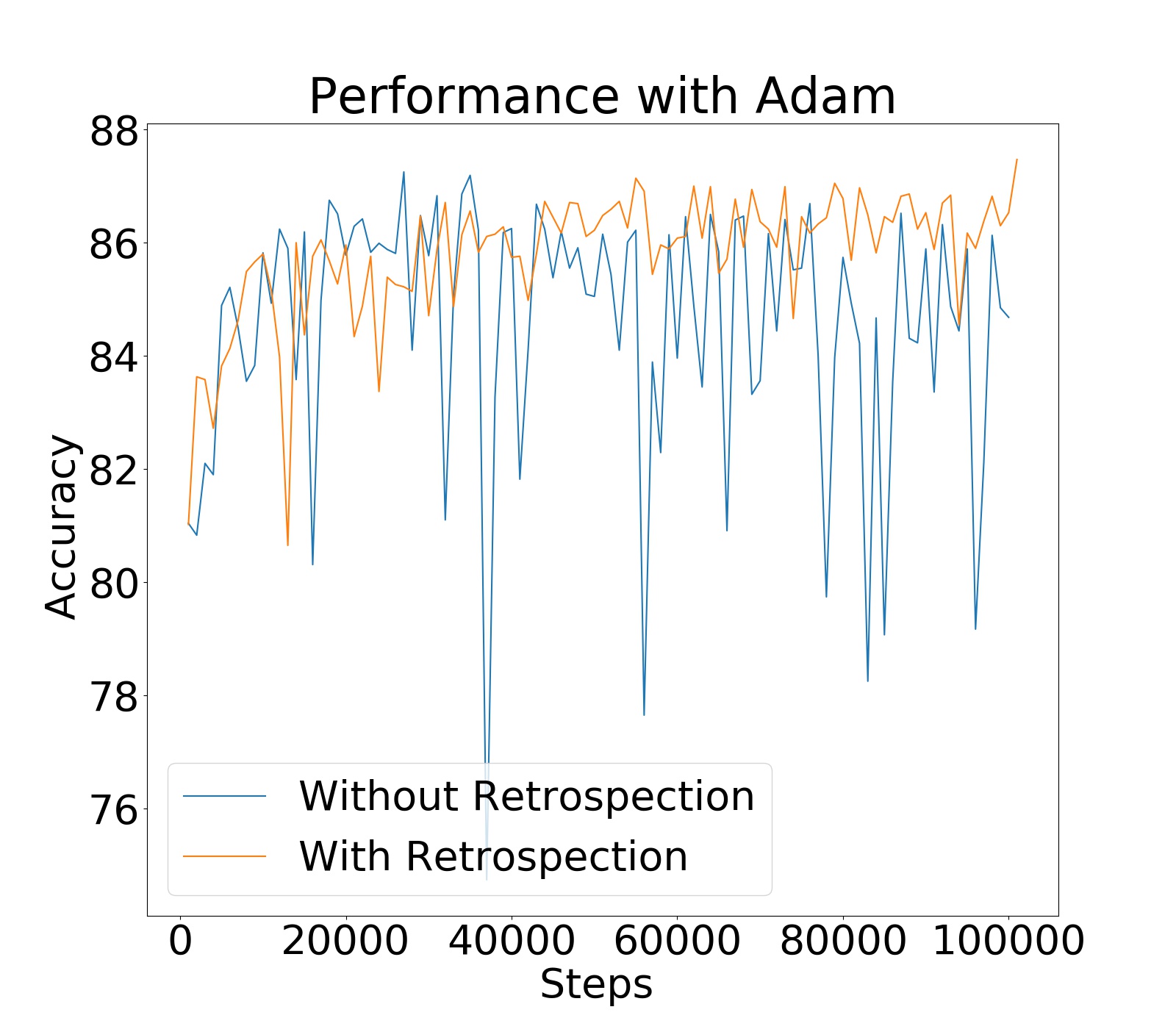}
  \end{minipage}
  \hspace{0.5cm}
  \begin{minipage}[b]{0.2\textwidth}
    \includegraphics[scale=0.11]{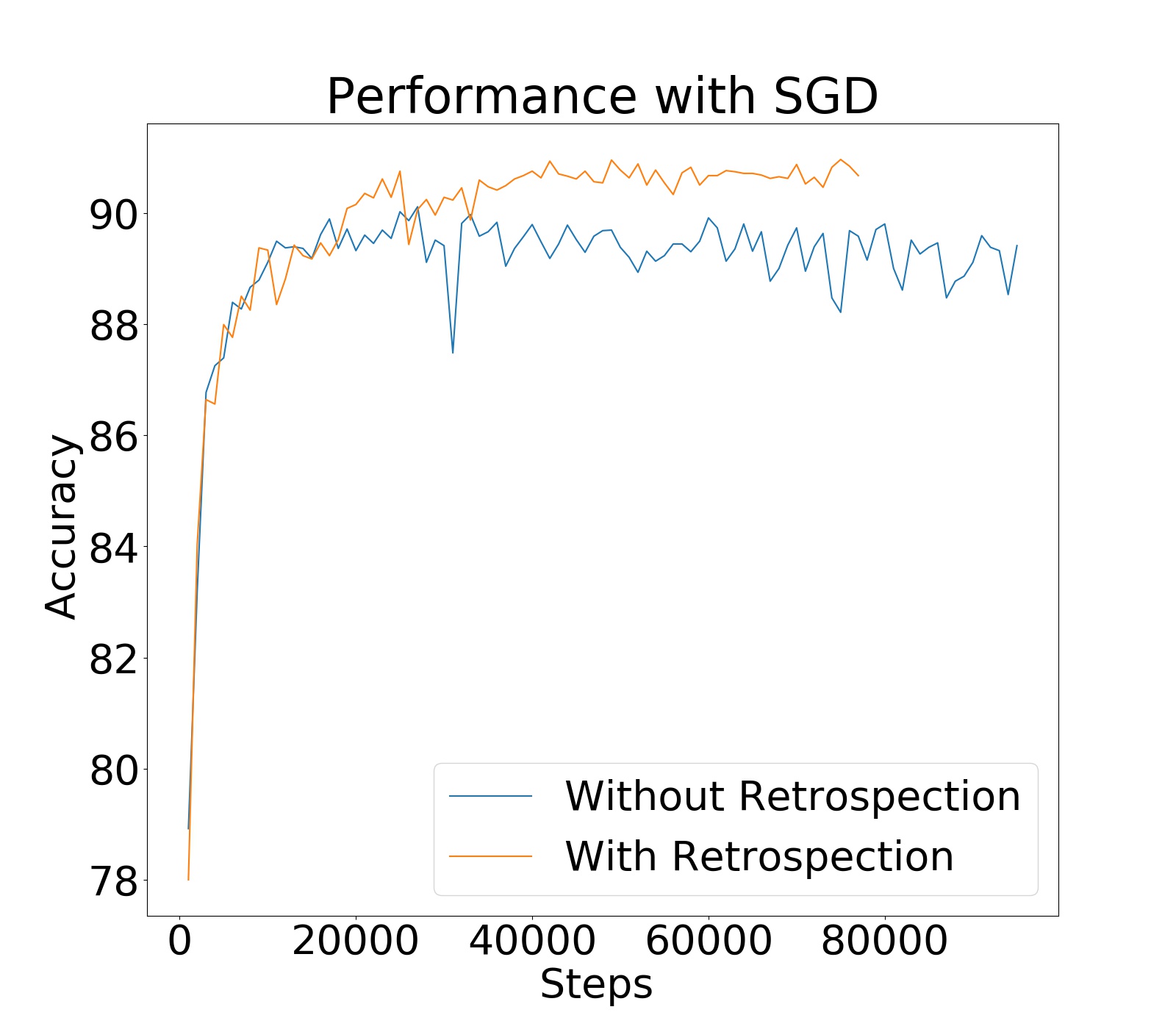}
  \end{minipage}
  \vspace{-4pt}
\caption{Image classification performance (test accuracy) using retrospective loss on LeNet with Adam and SGD (with momentum)}
\label{fig:ablation-optim}
\end{figure}

\noindent \textbf{Choice of Optimizer.}
We perform experiments to study the impact of the choice of optimizer to the proposed loss. We use Adam and SGD with momentum. The classification performance when using Adam and SGD (with momentum=0.5) are reported in Figure \ref{fig:ablation-optim}. While using retrospective loss improves performance in both cases, we notice that the improvement is more significant when using momentum. We ascribe this observation to the fact that retrospective loss and momentum use past model states/gradients in contrasting ways. Putting them together is perhaps a better approach, considering we derive the advantages of both methods.


This led us to additionally study the impact of the choice of momentum parameter in the same setting. We set momentum parameter to 0.5 in our corresponding experiments in Sec \ref{subsec_image_expts}. Here, we experimented with different values of the momentum parameter: (0.5, 0.7, 0.9). The results, reported in Table \ref{mom-ablation}, show that using retrospective loss always seems to add value to using momemtum, regardless of the momentum parameter, although the extent of improvement varies with the parameter choice. 

\begin{table}[h]
\centering
\begin{tabular}{c|c|c|c}
\hline
Config  & mom = 0.5  & mom = 0.7  & mom = 0.9   \\
\hline
Original      & 10.8 & 9.51 & 10.05 \\
Retrospective & 9.4  & 8.94 & 9.06 \\
\hline
\end{tabular}
\caption{Test error using LeNet on FMNIST using retrospection with different momentum (mom) parameter values.}
\vspace{-8pt}
\label{mom-ablation}
\end{table}

\begin{figure}[h]
\hspace{-2pt}
    \vspace{-8pt}
    \includegraphics[scale=0.1]{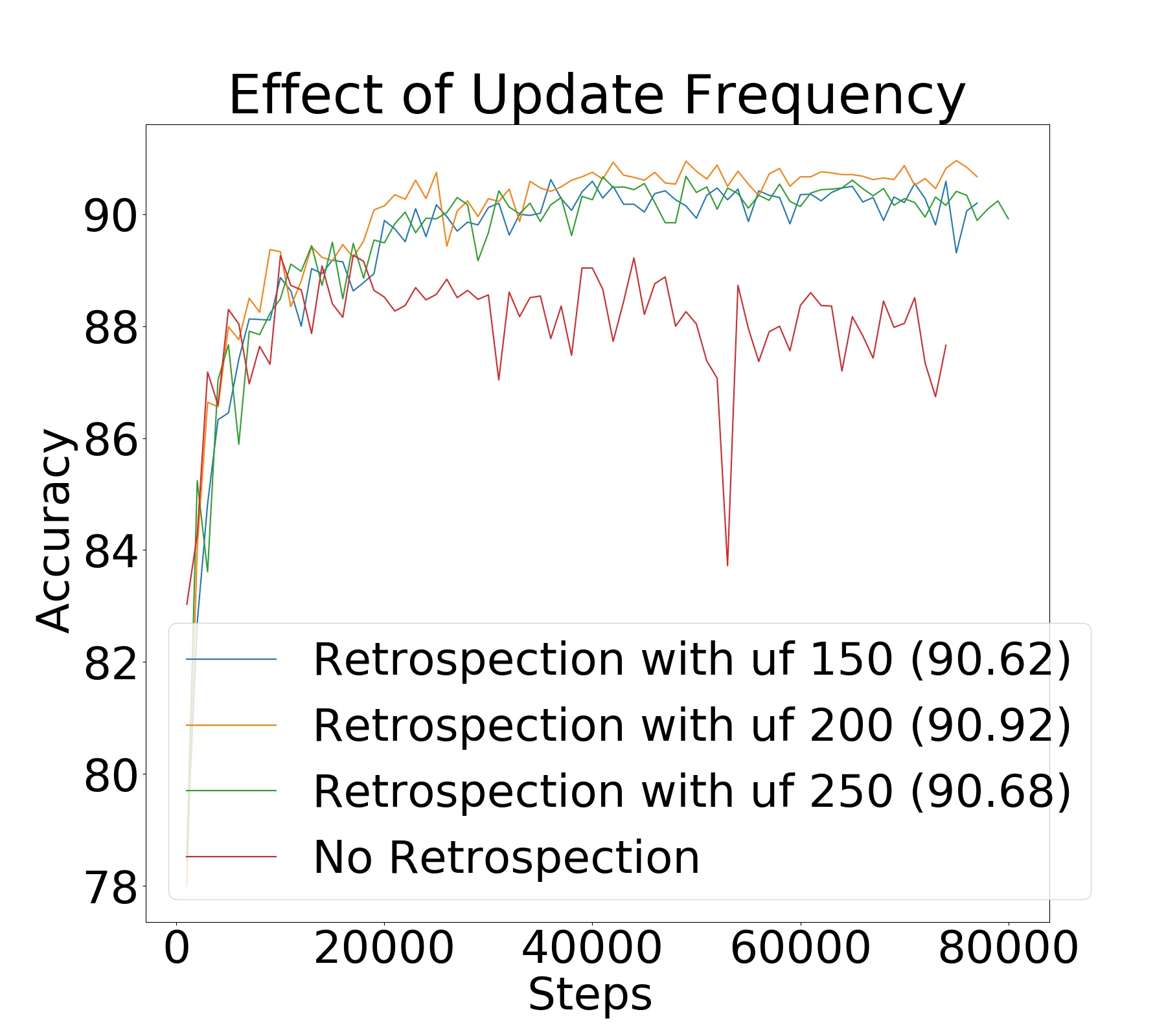}
    \vspace{-4pt}
\caption{Evolution of classification test accuracy when using different retrospective update frequencies}
\vspace{-8pt}
\label{fig:retro-update_freq}
\end{figure}

\noindent \textbf{Choice of Retrospective Update Frequency, $F$.} We study the impact of different update frequencies ($F$) for the retrospective loss. We experiment with 150, 200, 250 steps. Results are presented in Figure \ref{fig:retro-update_freq} with the best performance achieved using $F=250$ steps. Interestingly, all considered configurations of the retrospection loss outperform the configuration trained without it. We believe that mining for past parameter states to get maximal improvement in performance (training time or better model state) could be an interesting direction of future work. 

\noindent \textbf{Choice of Warm-up Period.}
We perform experiments to analyze the impact of choice of warm-up period. The error rates with different warm-ups are presented in Table \ref{warm-up-lenet}. We observed that on simpler datasets (like FMNIST), since networks start at a reasonable accuracy, retrospection is effective even when we introduce it with a very low warm-up period ($I_w$ = 0). The instability of the model during initial stages of training results in a slight dip in performance (although all choices of warm-up period are still better than the original loss alone).
\begin{table}[h]
\centering
\begin{tabular}{c|c|c|c|c|c}
\hline
Network         & Original                  & \multicolumn{4}{|c}{Retrospective} \\
\hline
                 & $I_{w}=0$   & $I_{w}=10k$   & $I_{w}=15k$   & $I_{w}=20k$   \\
\hline
LeNet & 10.05                              & 9.06            & 9.3              & 9.33             & 9.06 \\
\hline
\end{tabular}
\caption{Test classification error for different choices of warm-up period in retrospective loss.}
\vspace{-9pt}
\label{warm-up-lenet}
\end{table}

\begin{table}[]
\centering
\begin{tabular}{c|c|c|c}
\hline
Network   & Original & L1-norm & L2-norm \\
\hline
LeNet     & 10.8     & 9.4     & 9.7     \\
\hline
ResNet-20 & 7.6      & 6.8     & 7.3    \\
\hline
\end{tabular}
\caption{Test classification error using retrospective loss with different norms on F-MNIST}
\vspace{-10pt}
\label{fmnist-norm}
\end{table}
\noindent \textbf{Choice of Norm.} We analyzed the retrospective loss in Sec \ref{sec_methodology} and stated that $L_1$-norm maintains the consistency property, while $L_2$-norm does not. However, we study this empirically to judge the performance of $L_2$-norm version of the proposed loss. Table \ref{fmnist-norm} presents results of using retrospective loss with L1-norm and L2-norm. While both norms improve performance over training with the original loss alone, using L1-norm results in better performance, supporting our analysis in Sec \ref{sec_methodology}. Additional studies with other norms is another potential direction for future work.
\section{Conclusions and Future Work}
In this work, we introduced a new retrospective loss that utilizes parameter states from previous training steps to condition weight updates and guide the network towards the optimal parameter state. We presented the understanding of why it works, as well as conducted extensive experiments across multiple input domains, tasks, and architectures to show the effectiveness of the proposed loss across application domains. We also performed ablation studies on different hyperparameter choices which showed strong consistency in the usefuness of this loss in improving training of DNN models. The code for all our empirical studies with the
\begin{figure}[h]
\hspace{-2pt}
  \begin{minipage}[b]{0.2\textwidth}
    \includegraphics[scale=0.1]{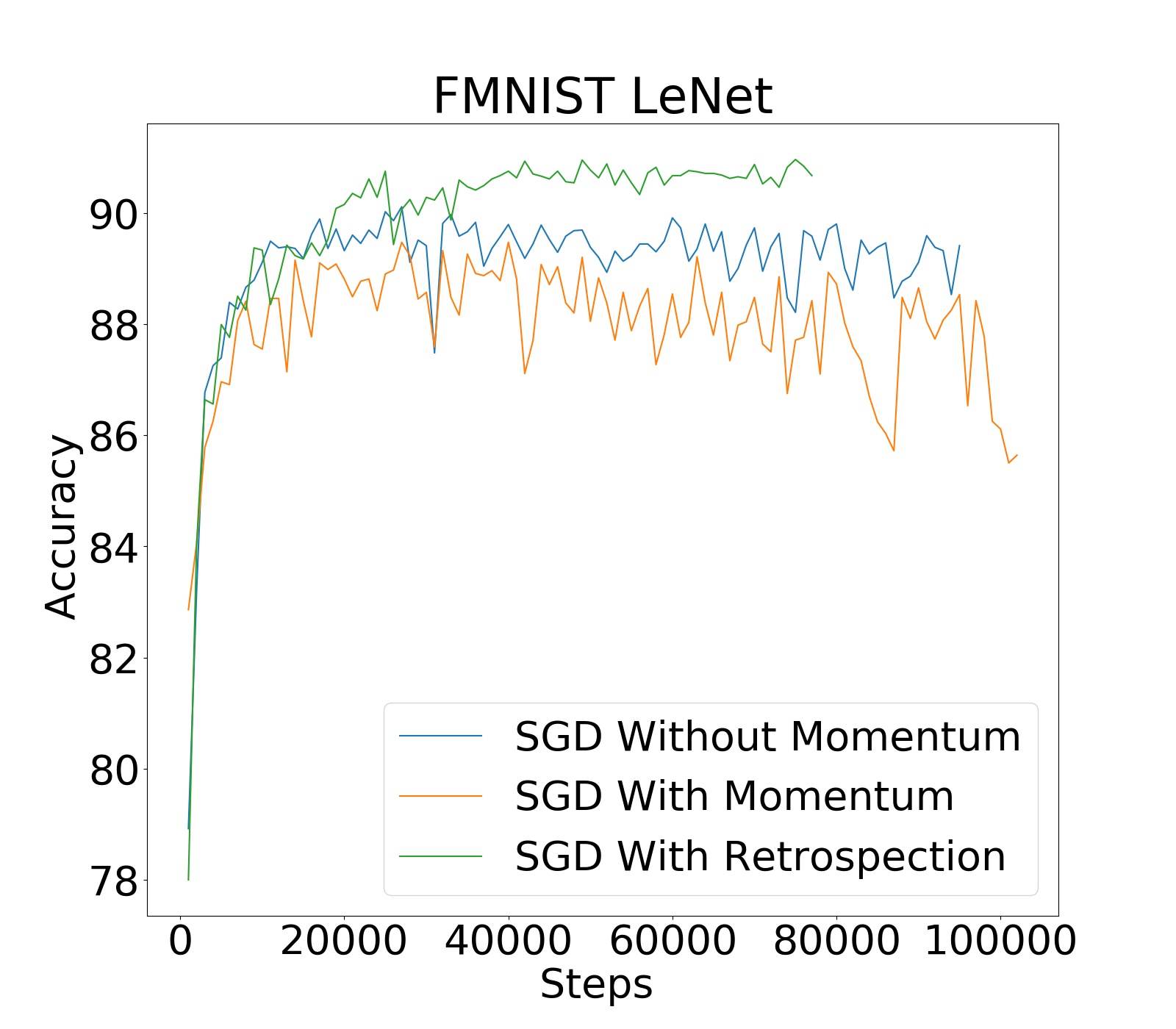}
  \end{minipage}
  \hspace{0.5cm}
  \begin{minipage}[b]{0.2\textwidth}
    \includegraphics[scale=0.1]{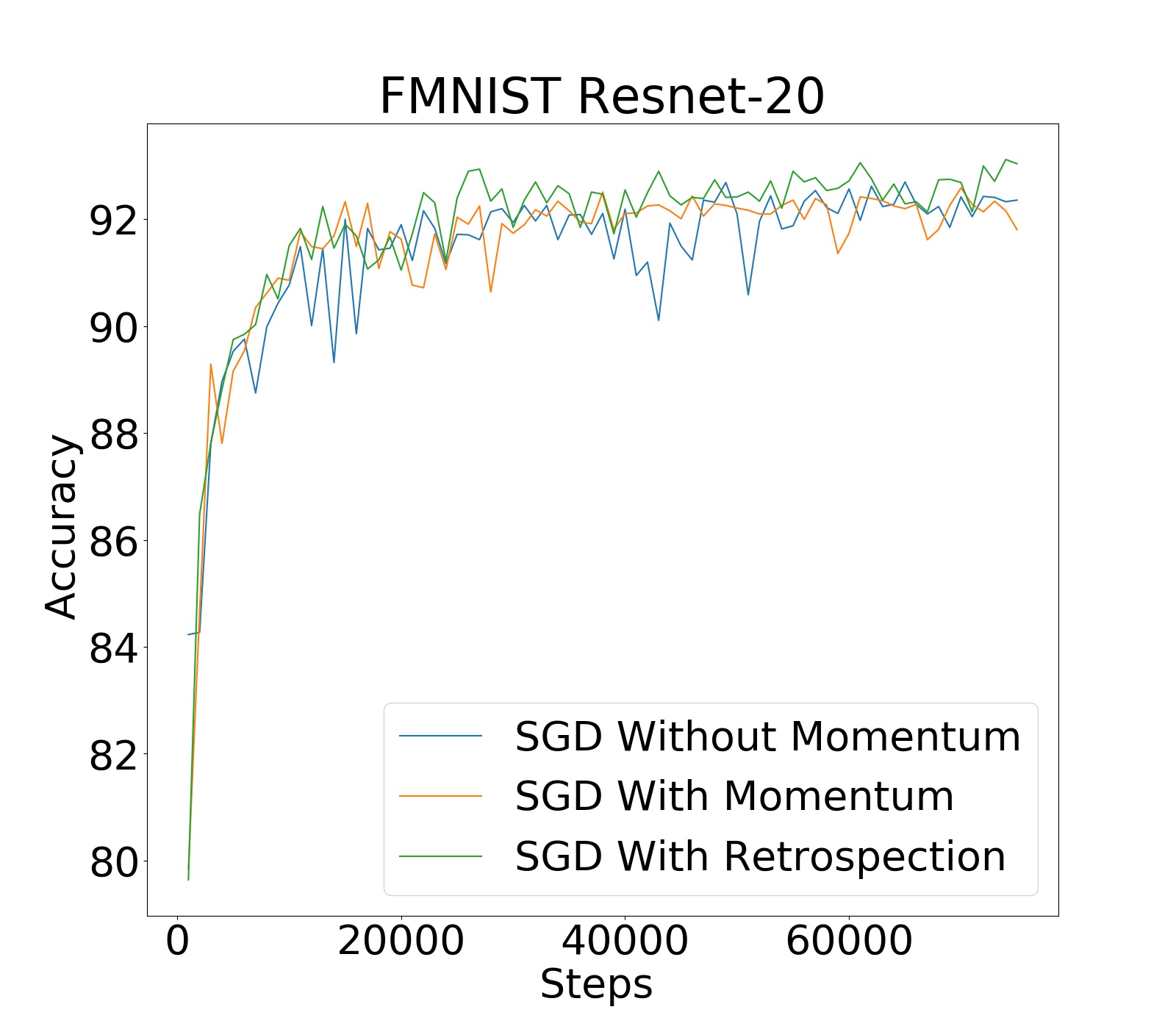}
  \end{minipage}
  \vspace{-5pt}
\caption{Comparison with momentum, LeNet on FMNIST}
\vspace{-10pt}
\label{fig:retro-momentum}
\end{figure}
Considering the common connection between retrospective loss and momentum for leveraging past model states/gradients although with different objectives, coupled with the results in Fig \ref{fig:ablation-optim}, we conducted further studies to understand this better. Figure \ref{fig:retro-momentum} shows initial results which compare performance from three configurations on image classification: (a) trained without retrospective loss (SGD); (b) trained without retrospective loss (SGD + momentum); and (c) with retrospective loss, without momentum (SGD + retrospection). The results show benefits of retrospective loss over momentum in this case, although one needs to carefully study this further to establish deeper connections. We consider this an important direction of our ongoing/future work, in addition to other pointers mentioned at different parts of the paper.

\bibliographystyle{ACM-Reference-Format}
\bibliography{kdd}
\end{document}